\documentclass[final,12pt]{clear2025} 

\usepackage{multirow}
\usepackage{multicol}
\usepackage{fancyhdr}
\usepackage{colortbl}
\usepackage{csquotes}
\usepackage{float}
\restylefloat{table}

\newcommand*{\affmark}[1][*]{\textsuperscript{#1}}

\title[Inducing Causal Structure Applied to Glucose Prediction for T1DM Patients]{Inducing Causal Structure for Interpretable Neural Networks Applied to Glucose Prediction for T1DM Patients}
\usepackage{times}
\usepackage{subcaption}

\clearauthor{%
 \Name{Ana {Esponera}}\affmark[1] \Email{anaesponera@gmail.com}\\
 \AND
 \Name{Giovanni {Cin{\`a}}}\affmark[1]\affmark[2] \Email{g.cina@amsterdamumc.nl}\\
 \\
\affmark[1] \addr Medical Informatics Dept., Amsterdam University Medical Center, The Netherlands\\
\affmark[2] \addr Institute for Logic, Language and Computation, University of Amsterdam, The Netherlands
}

\begin{document}

\maketitle

\begin{abstract}%
  Causal abstraction techniques such as Interchange Intervention Training (IIT) have been proposed to infuse neural network with expert knowledge encoded in causal models, but their application to real-world problems remains limited. This article explores the application of IIT in predicting blood glucose levels in Type 1 Diabetes Mellitus (T1DM) patients. The study utilizes an acyclic version of the \textit{simglucose} simulator approved by the FDA to train a Multi-Layer Perceptron (MLP) model, employing IIT to impose causal relationships. Results show that the model trained with IIT effectively abstracted the causal structure and outperformed the standard one in terms of predictive performance across different prediction horizons (PHs) post-meal. Furthermore, the breakdown of the counterfactual loss can be leveraged to explain which part of the causal mechanisms are more or less effectively captured by the model. These preliminary results suggest the potential of IIT in enhancing predictive models in healthcare by effectively complying with expert knowledge.

\end{abstract}


\section{Introduction}

Despite the great amount of machine learning models developed in both academia and industry, there is often uncertainty regarding their compliance with the underlying causal structures of the problems they aim to solve, which might be problematic in high-stake scenarios. This misalignment becomes particularly concerning when models are deployed in critical applications, such as medical prediction, where compliance with causal mechanisms is essential for reliability and safety. Recent research emphasizes the need for integrating causal reasoning into predictive models for decision support to enhance their robustness and interpretability 
\citep{vangeloven2024, Amsterdam2022,vanamsterdam2024}
. Causal models, specifically structural causal models (SCMs), play a pivotal role in this integration as they encode domain knowledge and illustrate how different variables influence one another. When the causal structure—represented as a DAG 
\citep{pearl2009causality}
—and the corresponding causal mechanisms in a SCM are known, causal abstraction provides mathematical tools to test whether a given machine learning model aligns with this SCM. Causal abstraction is defined as the alignment of a low-level model’s behaviour (e.g., a neural network) with the causal structure of a high-level model. By simplifying a complex system and focusing on the key causal relationships that drive outcomes, causal abstraction defines when a human-comprehensible, high-level causal explanation accurately represents opaque low-level details of a deep learning model \citep{DBLP:journals/corr/abs-2102-11107}. When this alignment is successful, models can reason more effectively, avoid spurious correlations, and comply with pre-existing knowledge.

An essential aspect of causal reasoning is counterfactual reasoning, which involves evaluating \enquote{what-if} scenarios by manipulating specific variables and observing the resulting changes in outcomes.
IIT is a training method based on counterfactual behaviour optimization, designed to impose causal structures on neural networks \citep{Geiger2022}. While this technique was demonstrated on simple problems such as visual recognition and natural language processing tasks, it remains to be shown whether it can make a difference in a real-world setting.

One such use cases is the management of T1DM. T1DM is a prevalent chronic metabolic condition characterized by the insufficient absorption of glucose by the body's cells, leading to elevated levels of blood glucose (BG) \citep{ijms22147644,sref, Aloke2022}. This condition results from the lack of insulin production, requiring external insulin administration which affects 8.75 million patients worldwide \citep{IDF_T1D_Index_Report_2022} and is estimated to cost 966 billion U.S. dollars globally in healthcare expenditures \citep{Statista_Diabetes_Costs_2024}. BG prediction poses unique challenges due to the complex and dynamic nature of glucose metabolism, influenced by numerous factors such as food intake, physical activity, and insulin administration. Accurately predicting BG levels is essential for effective T1DM management, yet existing simulators may be computationally intensive and unsuitable for integration into lightweight, wearable medical devices, such as insulin pumps \citep{s24196322, NAHAVANDI2022106541}. A neural network that can serve as a computationally efficient, causal abstraction of BG prediction could enable real-time, on-device predictions, potentially improving T1DM management and reducing healthcare burdens.

In this paper, we apply IIT to the prediction of BG in T1DM patients. Our primary objective is to demonstrate the applicability of IIT in a real-world context, specifically in enhancing predictive models within the healthcare domain by leveraging expert knowledge encoded in causal models. We detail the improvements of IIT-trained models over standard models in terms of (a) model performance, (b) data efficiency, and (c) compliance with expert knowledge. Finally, we discuss the challenges encountered when integrating this technology into real-world healthcare scenarios.

\section{Preliminaries}
Interchange intervention training is a technique designed to inject causal structure into neural network models (NN) by leveraging counterfactual reasoning.
The goal is to guide the model towards more robust learning by focusing on the underlying causal logic of the task, rather than merely capturing surface correlations.
To this end, it is essential to establish a mapping between the causal DAG and the architecture of the neural network. This mapping specifies which parts of the neural network correspond to specific nodes in the DAG, effectively assigning roles within the network to components of the causal model. Different mappings can yield different results, as they influence how the high-level causal structure is embedded within the low-level computations of the NN.
Exploiting this mapping, an interchange intervention involves swapping part of
the input value while the rest stays constant, and then
comparing the outcomes under both scenarios.
~\citet{Geiger2022} defines an interchange intervention as a model used to process two different inputs ($source$ and $base$) and then a particular internal state obtained by processing $source$ is used in place of the corresponding internal state obtained by $base$.
This allows researchers to estimate the counterfactual effect of that part of the intervened value by effectively holding other variables constant or \enquote{interchanging} them.
The process establishes and aligns a proposed causal structure with the model.
The alignments ensure that (clusters of) lower-level variables accurately reflect or capture the high-level variables in the SCM.
Importantly, these interventions are not limited to the input layer but can be applied to any intermediate value within the causal structure. The loss in such counterfactual scenario is dubbed $L_{INT}$ \citep{geiger2024findingalignmentsinterpretablecausal}: 
\begin{small}
\begin{equation} \label{intervention_loss_eq}
L_{INT} = \sum_{b,s\in in}^{} Loss(M_{H_{\tau(I\xleftarrow{}b,s)}}, M_{L_{I\xleftarrow{}b,s}})
\end{equation}
\end{small}
where $b$ and $s$ are the $base$ and $source$ input values from the $in$ input space swapped during the intervention, $I$ is the variable being intervened on, $I\xleftarrow{}b,s$ is the intervention mechanism, $M_{H}$ is the high-level neural model, $M_{L}$ is the low-level causal model, $\tau$ is a mapping of output values from the low to high-level and $Loss$ is the chosen function to quantify the distance.
If the $L_{INT}$ is reduced to zero, we can guarantee that the target causal model serves as a causal abstraction of the neural model \citep{Geiger2022}. Crucially, this procedure does not guarantees correctness when the starting knowledge is flawed: if the SCM is encoding incorrect knowledge then IIT will align the low-level model to the ill-specified SCM.
By training the model with IIT and encouraging it to reduce $L_{INT}$, the desired causal structure can be effectively imposed on a neural network. Since interventions are local, i.e. they pertain to a subpart of the SCM, IIT also allows for a breakdown of which components of the SCM are successfully abstracted by the NN, adding to the interpretability of the model.

\section{Related work}\label{related-work}

\citet{Geiger2022} introduce IIT and evaluate it across three tasks: a structural vision task (MNIST-PVR), a navigational language task (ReaSCAN), and a natural language inference task (MQNLI). They demonstrate that IIT outperforms multi-task training and data augmentation, yielding more interpretable neural models aligned with the intended causal structure. 
In the context of model distillation, \citet{distillation-iit} apply IIT to BERT, achieving significant improvements. Specifically, IIT reduces perplexity on masked language modeling tasks and enhances performance on benchmarks such as GLUE, SQuAD, and CoNLL-2003.
Furthermore, \citet{huang2023inducingcharacterlevelstructuresubwordbased} propose Type-level Interchange Intervention Training (TIIT) to induce character-level structures in subword-based models, improving robustness and the model's understanding of subword dependencies.
As the most recent application of IIT, \citet{gupta2024interpbenchsemisynthetictransformersevaluating} present a framework that leverages IIT with known circuits for evaluating mechanistic interpretability methods. Their experimental results demonstrate that this approach not only enhances performance on challenging benchmarks but also provides deeper insights into the internal mechanisms driving model predictions.

In the field of machine learning for BG prediction in T1DM patients, \citep{metanylis} provide a meta-analysis showing that NN exhibit the highest performance across different prediction horizons (PH). Regarding models that incorporate glucose-insulin dynamics, we have specifically chosen to focus on the works of \citet{Rebaz_EGA_benchmark} and \citet{Lui_EGA_benchmark} because both report clear metrics that allow for direct comparison, employ the same PHs and comply with GLYFE benchmark \citep{benchmark_GLYFE_2021}. GLYFE establishes a reproducible benchmark for evaluating diverse machine learning models for personalized glucose forecasting in T1DM across both simulated, namely UVA/Padova Type 1 Diabetes Metabolic Simulator, and real patient dataset Ohio Type-1 Diabetes Mellitus. \citet{Rebaz_EGA_benchmark} introduce a novel BG prediction method using an absorption model, achieving superior performance for 60- and 120-minute PHs. Finally, \citet{Lui_EGA_benchmark} present a deconvolutional model based on glucose-insulin dynamics, outperforming conventional models for PHs beyond 60 minutes.

\section{Data and Methods}
\subsection{Experimental set up and metrics}\label{sec:experimental_setup}
The predictions were evaluated on the 30, 45, 60 and 120-minute PH, following the patients' first meal of the day, standardized as breakfast. Patients were assumed not to consume any extra food between this meal and the subsequent glucose measurement, and were all treated with a specific dose of insulin. Glucose levels, which are continuous measurements typically ranging from 40 to 400 mg/dL in diabetic patients, were monitored throughout the simulation. The performance metrics used in the evaluation are: average absolute error (MAE), mean squared error (MSE), root mean squared error (RMSE). Since not all prediction errors are the same when it comes to vital signs, we included as metric the percentage of predicted values in the \enquote{clinically acceptable} EGA classes A and B \citep{ega-paper}. EGA grids can be used as an indicator of underfitting in a model, for instance when the predictions display a horizontal trend. Detailed information regarding the training configuration and evaluation metrics is reported in Appendix~\ref{ap:experimental_setup}. 

\subsection{Data acquisition, preprocessing and setting}\label{sec:data_acquisition}
The data used in this study is based on in-silico data from \citet{uva_padova}. The entire dataset consists of 30 T1DM patients (10 children, 10 adolescents, 10 adults) with characteristics derived from three joint parameter distributions. Each virtual patient is described by the patient's initial state $x$, multiple kinetics constants $k$, an action representing carbohydrate intake $CHO$ and insulin dosage $insulin$ (see Fig.~\ref{fig:simglucose}). In total, 64 parameters are provided per patient as a subject model parameter vector. As the provided number of patients is limited, we generate a specific number of observations comparable to existing cohorts used in previous studies in the literature \citep{Lui_EGA_benchmark}. Therefore, 200 T1DM subjects were generated using Gaussian distributions conditioned by age from the FDA-approved in-silico joint distributions (7 children, 10 adolescents, 183 adults). More details about UVA/Padova T1DM joint distributions can be found in \citet{uva_padova, uva_padova_2} and more details about the data derivation can be found in Appendix \ref{ap:data_acquisition}.

Out of those 64 parameters, only 9 were selected for input to the model. The parameter vector consists of 51 pharmacokinetic constants and 13 patient state variables at the initial time $t_{n}$. The 51 pharmacokinetic constants were excluded from the input dataset because they are fixed for each patient and our objective is to develop a model that generalizes across patients without this fine-grained information. While the 13 patient state variables are patient-specific, they represent dynamic states that could in principle be available for the model. In practice, the model is only fed with the information that can reasonably be measured, allowing it to infer the remainder. Four patient state parameters, $x_1$, $x_2$, $x_3$, and $x_7$, were also excluded as they were initialized to 0 for all patients. They represent the amount of glucose and insulin at the initial time $t_{n}$ which is negligible for the breakfast scenario. The final input data comprised 20 parameters: 9 from the patient's initial state, 9 from pre-meal glucose levels, and 2 for insulin and CHO intake. Preprocessing included Z score standardization for patient's state parameters, and glucose levels were scaled at a 100:1 ratio to improve training stability without distorting relationships between values. The data was split 80/20, with 160 patients in the training set and 40 in the validation set. The test set only included the original 30 FDA-approved in-silico patients to evaluate model generalization to FDA-approved data.

\subsection{The \textit{ Simglucose} simulator\label{sec:simglucose}}

To use IIT, a causal model is required. We selected the UVA/Padova T1DM simulator, \textit{simglucose}, a collaborative effort between the University of Virginia and the University of Padova. It simulates physiological processes and glucose-insulin dynamics in T1DM patients, serving as a substitute for preclinical trials of insulin treatments, including closed-loop algorithms.

Three versions of the simulator have been released so far: S2008, S2013 and S2017. Each version incorporates an improved modelling of glucose-insulin interactions.  \citet{s2008vss2013} detected that the incidence of hypoglycemic events projected by the S2008 simulator did not entirely align with those recorded in clinical trials. Therefore, the S2013 version incorporated a new insulin-dependent compartment that improves the model of glucose kinetics in hypoglycemia. Later on, the S2017 release improved the intra-day glucose variability and the nocturnal glucose increase. The three versions have been validated and accepted by FDA. However, S2013 and S2017 official implementations in \textit{MatLab} do not have an academic license. For this reason, we have chosen the S2008 version, which is open source for Matlab and Python and is summarized in Fig~\ref{fig:simglucose}. Beside the limitation in hypoglycemic scenarios, the S2008 simulator faithfully represents glucose dynamics in T1DM patients for euglycemic and hyperglycemic scenarios.
More details on the model are available in \citet{uva_padova, uva_padova_2}. At its full complexity, the \textit{simglucose} causal model for $PH_{n}$ corresponds to a DAG unfolding through n steps in time. In order to obtain alignment with a non-recurrent NN - which effectively predicts only the BG at the PH in one leap - we elected to simplify this DAG compressing the time dimension. This operation produced two cyclic relationships between equations $dx_{4}dt$/$dx_{5}dt$ and $dx_{6}dt$/$dx_{10}dt$; acyclicity was recovered by removing the terms colored in red in the figure. We refer to this as the \enquote{amended} \textit{simglucose}.
The decision to experiment with a time-compressed causal model was driven by the need for simplicity and the fact that the structural equations themselves are stable over time. While this approach may not fully capture temporal dependencies, it provides a tractable starting point for applying IIT in a healthcare setting.

\begin{figure}[h!]
\caption{\textit{Simglucose} specific kinetic constants and details on the model}
\label{fig:simglucose}
\centering
\fbox{%
\begin{minipage}{1\linewidth}
\smaller
\begin{multicols}{2}
\hspace*{0.75cm}\textbf{Equations:}\begin{align*}
dx_{1}dt & = -k_{\text{max}} \cdot x_1 + \text{CHO} \\
dx_{2}dt & = k_{\text{max}} \cdot x_1 - x_2 \cdot k_{\text{gut}} \\
dx_{3}dt & = k_{\text{gut}} \cdot x_2 - k_{\text{abs}} \cdot x_3 \\
dx_{4}dt & = EGP_t + Ra_t - U_{\text{iit}} - E_t - k_1 \cdot x_4 \\
       & \quad + k_2 \cdot x_5 \\
dx_{5}dt & = -U_{\text{idt}} \,{\color{red} + k_1 \cdot x_4} - k_2 \cdot x_5 \\
dx_{6}dt & = -(k_{\text{m2}} + k_{\text{m4}}) \cdot x_6 + k_{\text{m1}} \cdot x_{10} \\
       & \quad + k_{\text{a1}} \cdot x_{11} + k_{\text{a2}} \cdot x_{12} \\
dx_{7}dt & = -k_{\text{p2u}} \cdot x_7 + k_{\text{p2u}} \cdot (I_t - k_{\text{Ib}}) \\
dx_{8}dt & = -k_i \cdot (x_8 - I_t) \\
dx_{9}dt & = -k_i \cdot (x_9 - x_8) \\
dx_{10}dt & = -(k_{\text{m1}} + k_{\text{m3}}) \cdot x_{10}\, {\color{red} + k_{\text{m2}} \cdot x_6 }\\
dx_{11}dt & = \text{action}_{\text{insulin}} - (k_{\text{a1}} + k_d) \cdot x_{11} \\
dx_{12}dt & = k_d \cdot x_{11} - k_{\text{a2}} \cdot x_{12} \\
dx_{13}dt & = -k_{\text{sc}} \cdot x_{13} + k_{\text{sc}} \cdot x_4
\end{align*}

\columnbreak

\noindent
\textbf{Where:}
\begin{align*}
x_1 &: \text{Amount of glucose in the stomach (solid-state) (mg)} \\
x_2 &: \text{Amount of glucose in the stomach (liquid state) (mg)} \\
x_3 &: \text{Glucose mass (GM) in the intestine (mg)} \\
x_4 &: \text{GM in plasma and rapidly equilibrating} \\
& \quad \text{tissues (ET) (mg/kg)} \\
x_5 &: \text{GM in tissue and slowly ET (mg/kg)} \\
x_6 &: \text{Insulin mass in plasma (pmol/kg)} \\
x_7 &: \text{Insulin in the interstitial fluid (pmol/L)} \\
x_8 &: \text{Plasma insulin concentration (pmol/L)} \\
x_9 &: \text{Delayed insulin (pmol/L)} \\
x_{10} &: \text{Insulin mass in the liver (pmol/kg)} \\
x_{11} &: \text{Nonmonomeric insulin in subcutaneous}\\
& \quad \text{ subcutaneous space (pmol/kg)} \\
x_{12} &: \text{Monomeric insulin in subcutaneous space (pmol/kg)} \\
x_{13} &: \text{Subcutaneous glucose level (mg/kg)}
\end{align*}
\end{multicols}
\begin{multicols}{2}
\hspace*{0.75cm}\textbf{Model Variables:}
\begin{align*}
U_{\text{idt}} &: \text{insulin-dependent glucose} \\
& \quad \text{ utilization (mg/kg/min)} \\
EGP_t &: \text{Endogenous glucose} \\
& \quad \text{ production (mg/kg/min)} \\
Ra_t &: \text{Glucose rate of appearance} \\
& \quad \text{ in plasma (pmol/kg/min)} \\
U_{\text{iit}} &: \text{Insulin-independent glucose} \\
& \quad \text{ utilization (mg/kg/min)} \\
E_t &: \text{Renal excretion (mg/kg/min)} \\
I_t &: \text{Plasma insulin concentration (pmol/liter)} \\
k_{\text{sc}} &: \text{Amount of nonmonomeric and monomeric} \\
& \quad \text{ insulin in the subcutaneous space (pmol/kg)} \\
k_{\text{Ib}} &: \text{Plasma insulin concentration} \\
& \quad \text{at basal state (pmol/liter)} \\
\end{align*}

\columnbreak

\noindent
\textbf{Constants: ($min^{-1}$)}
\begin{align*}
k_{\text{max}} &: \text{Rate of grinding} \\
k_{\text{gut}} &: \text{Rate gastric emptying} \\
k_{\text{abs}} &: \text{Rate intestinal absorption} \\
k_{\text{p2u}} &: \text{Rate insulin action on peripheral} \\
& \quad \text{ glucose utilization} \\
k_i &: \text{Rate parameter accounting for delay between} \\
& \quad \text{ insulin signal and insulin action} \\
k_d &: \text{Rate insulin dissociation} \\
k_1, k_2 &: \text{Rate parameter of glucose kinetics} \\
k_{\text{m1}}, k_{\text{m2}},  &: \text{Rate parameter of insulin kinetics} \\
 k_{\text{m3}}, k_{\text{m4}} &: \text{Rate parameter of insulin kinetics} \\
k_{\text{a1}} &: \text{Rate nonmonomeric insulin absorption} \\
k_{\text{a2}} &: \text{Rate monomeric insulin absorption} \\
\end{align*}
\end{multicols}
\normalsize
\end{minipage}
}
\end{figure}

\subsection{NN and alignment to causal model}\label{sec:teacher_student}
\begin{figure}
    \centering
    \subfigure[]{\includegraphics[width=0.40\textwidth]{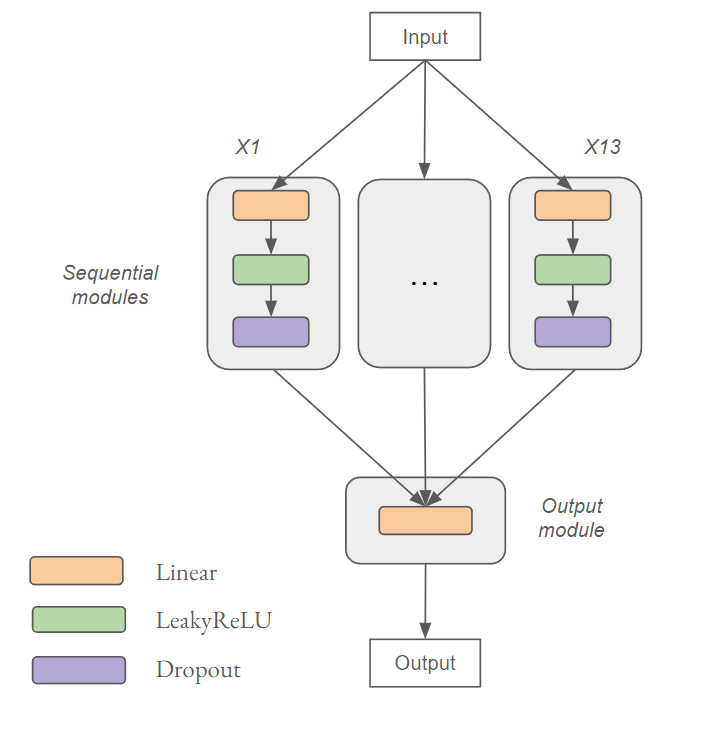}\label{fig:nn-models-1}} 
    \subfigure[]{\includegraphics[width=0.32\textwidth]{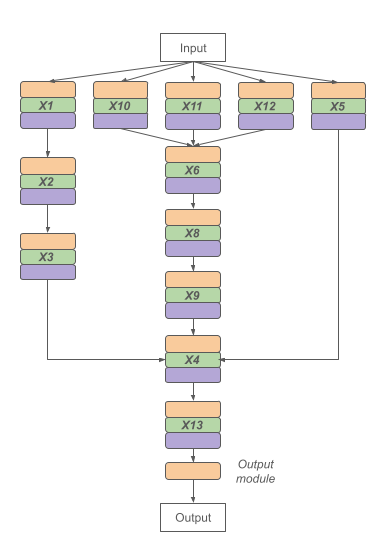}\label{fig:nn-models-2}} 
    \subfigure[]{\includegraphics[width=0.26\textwidth]{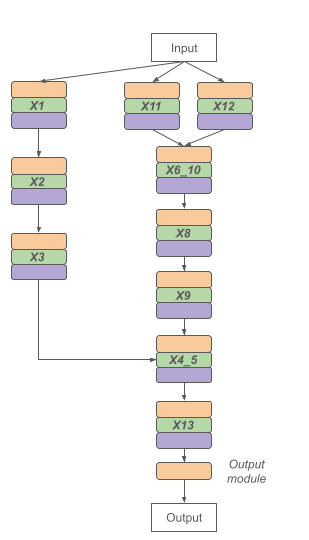}\label{fig:nn-models-3}}
    \caption{NN architecture diagram for (a) MLP parallel, (b) MLP tree, and (c) MLP joint}
    \label{fig:nn-models}
\end{figure}

We use a minimal building block to approximate each node of the DAG representing the causal structure. Each building block consists of a linear layer followed by a leaky ReLU activation function and a dropout layer with a rate of 0.3 (See Fig.~\ref{fig:nn-models}). By assembling these building blocks, we construct neural network models that approximate the causal structure at varying complexity levels. 

Our main model is the \textbf{MLP tree architecture}.
The architecture of this model establishes connections between the 13 sequential modules, imitating the blueprint of the causal model (See Fig~\ref{fig:nn-models-1}). Each sequential module corresponds to one of the patient state parameters from $x_1$ to $x_{13}$. For this NN, two versions of this architecture exist: one with $hidden\_size$ 128 and another with more flexibility of $hidden\_size$ 256.

In this model, each sequential module approximates the function that computes a variable in the structural causal model: variable $x_1$ from the computational model is aligned to the sequential module $X1$ in the NN, and so on. Because of the compression of the DAG on the time dimension, each module's output corresponds to the value of the corresponding causal variable at the prediction horizon after the meal.

To assess the effect of alternative architectures and the impact of the aforementioned operations on the DAG, we also consider two variations.

The first one is \textbf{MLP parallel architecture}. It is composed of 13 identical sequential modules ($X1$ to $X13$) arranged in parallel, with no connections between them (See Fig~\ref{fig:nn-models-2}). This model does not capture the causal dependencies among the variables and serves as a baseline to evaluate the importance of incorporating the causal structure. The alignment follows the same alignment as in the MLP tree model.

The second one is \textbf{MLP joint architecture} . It addresses the cyclic connections in the compressed DAG by merging the related modules ($dx_{4}dt$ with $dx_{5}dt$, and $dx_{6}dt$ with $dx_{10}dt$) into joint modules (See Fig~\ref{fig:nn-models-3}). This model can therefore align with the model described in Figure \ref{fig:simglucose} including the red component, albeit not with an injective mapping. The latter is adjusted as follows: the joint sequential module $X4\_5$ is aligned to both $x_4$ and $x_5$, and $X6\_10$ is aligned to both $x_6$ and $x_{10}$. For the rest of the sequential modules, the alignment follows the same alignment as before.

\section{Results}
\subsection{Performance of MLP tree with and without IIT}
\begin{figure}[h]
    \centering
    \includegraphics[scale=0.6]{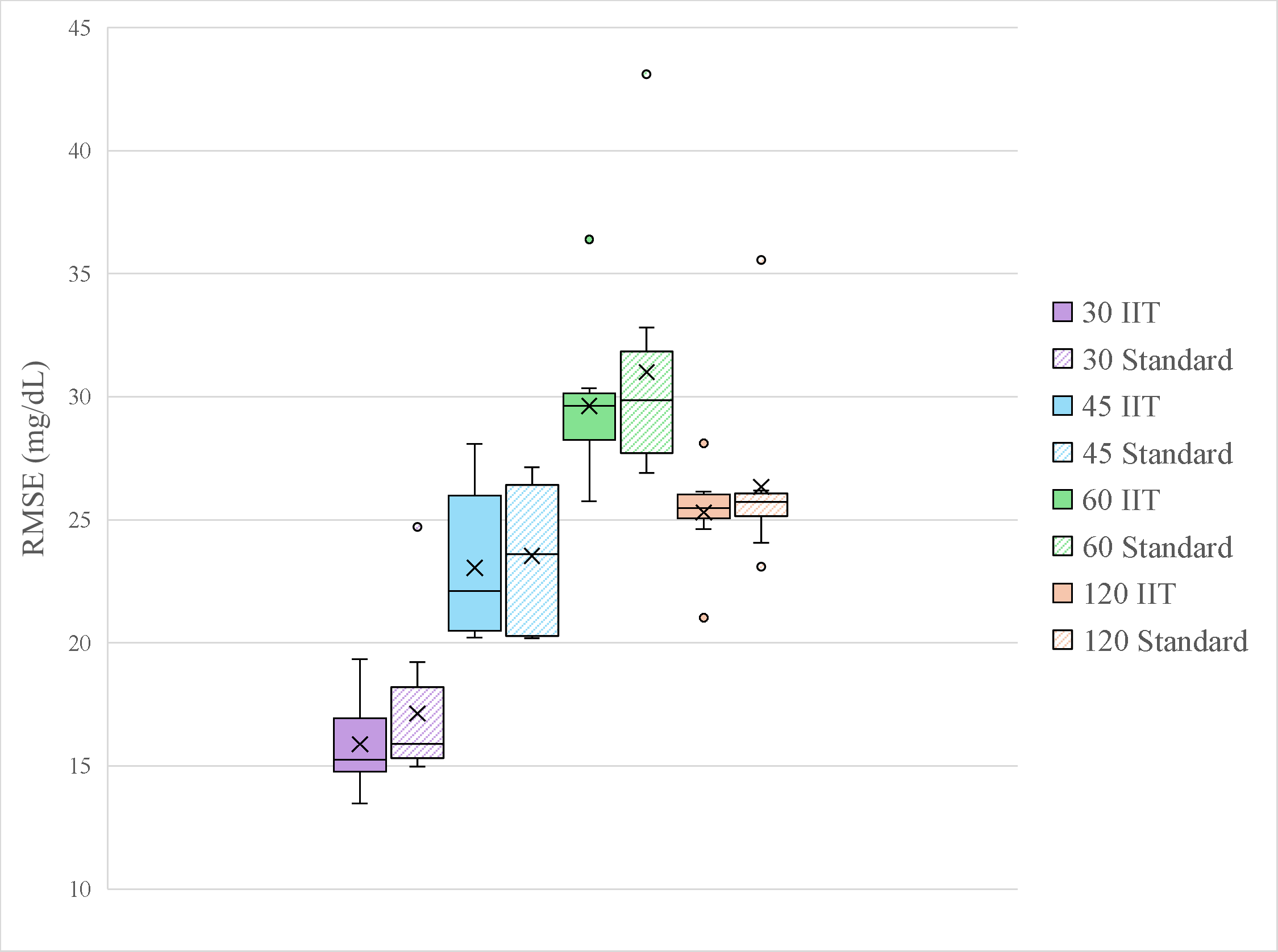}
    \caption{RMSE (mg/dL) prediction error of the model MLP tree (256) IIT using amended \textit{simglucose} across the four prediction horizons (PH) for the test (n=30) in-silico T1DM patients and 10 different random seeds. The solid bars refer to IIT trainings while the striped bars refer to standard training. The mean is indicated by the cross.}
    \label{fig:results_iit_modified_vs_regular}
\end{figure}

The box plot in Figure~\ref{fig:results_iit_modified_vs_regular} compares RMSE across PHs for the MLP tree model with 256 hidden units, using amended \textit{simglucose} as causal model. Solid-coloured columns represent IIT-trained model, and trace-filled columns represent standard-trained model. Lower RMSE values indicate better predictions. Overall, IIT training consistently achieves lower RMSE values than Standard training across all PHs, highlighting its effectiveness. The median RMSE values are visibly lower for IIT models. This is particularly pronounced at the 30-minute and 45-minute PHs, where IIT training shows significant reductions in RMSE (approximately 16 and 23 mg/dL, respectively), compared to the standard training counterparts, which have higher medians and a wider distribution of errors.
At 60 and 120-minute PHs, the RMSE differences between IIT and standard models are less pronounced, but IIT-trained models still maintain a lower error trend. The interquartile ranges are narrower for IIT training, indicating more consistent model performance across multiple runs, whereas standard training tends to have wider distributions, implying greater variability.
Detailed values for MSE, MAE and EGA performance are reported in Appendix~\ref{table:cyclic}. Similarly, the IIT-trained model achieves lower MSE, lower MAE and higher EGA across all PHs compared to standard-trained model. The difference is substantial for PH 60 (MSE reduction = -95.89 $(mg/dL)^2$) and 120 (MSE reduction = -128.34 $(mg/dL)^2$).
In addition, Appendix~\ref{fig:ega} shows Clarke error grid analysis to visualize predictions within clinically accepted ranges. Only the representative 120 PH grid is shown. The model MLP tree with 256 as hidden size Fig.~\ref{subfig:tree-256-sim-iit},~\ref{subfig:tree-256-sim-standard} shows dispersed predictions along the diagonal for both IIT-trained and standard-trained model.

\subsection{Causal abstraction analysis}
\begin{figure}[h]
    \centering
    \subfigure[]{\includegraphics[width=0.45\textwidth]{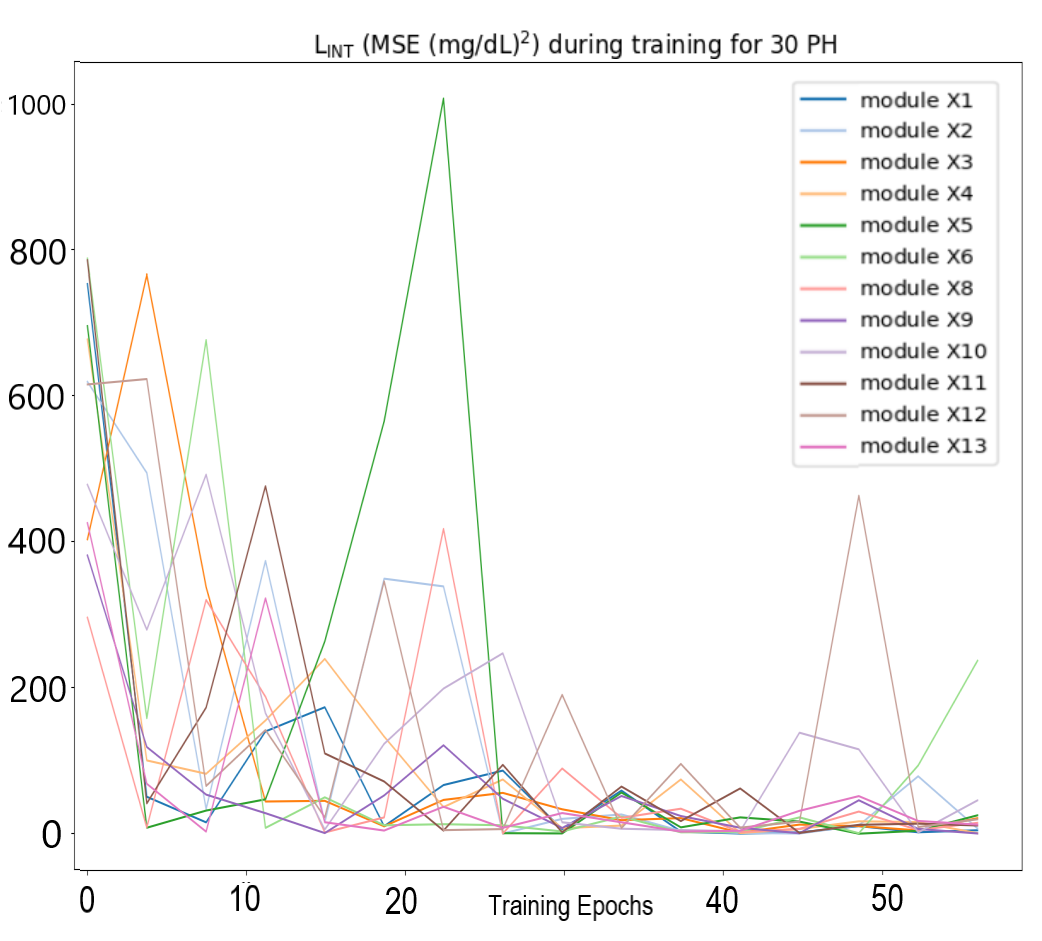}\label{fig:results_loss_int_fig_short}}
    \subfigure[]{\includegraphics[width=0.52\textwidth]{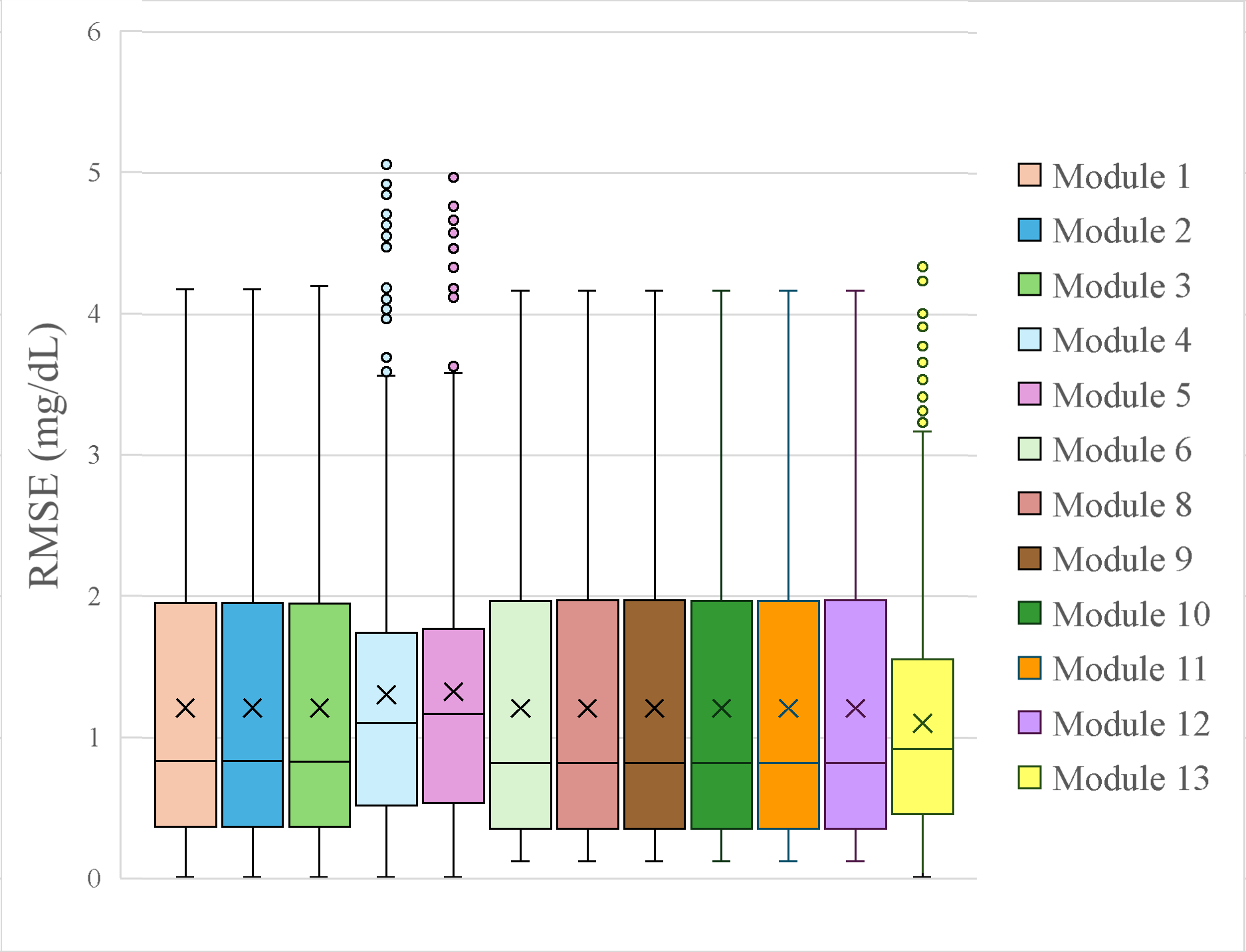}\label{fig:results_loss_test_int_figs_short}}
    \caption{$L_{INT}$ tracking for the MLP tree model for PH 30. The $L_{INT}$ is grouped by modules. a) during training and b) during testing.}
\end{figure}

We tracked the counterfactual loss ($L_{INT}$) during the training process across different PHs for the MLP tree model with 256 hidden units. Figure~\ref{fig:results_loss_int_fig_short} displays the results for 30 PH, showing that $L_{INT}$ consistently decreases as training epochs progress; the same holds for the other PHs. The corresponding plots are included in Appendix~\ref{sec:causal-abst} Fig~\ref{fig:results_loss_int_fig}.
With more detail, Figure~\ref{fig:results_loss_test_int_figs_short} visualizes the locality, spread, and skewness of $L_{INT}$ for the 30 PH to facilitate analysis of causal abstraction per module. Module $X4$ and module $X5$ tend to have a slightly higher $L_{INT}$ median than the rest of the modules. Modules $X4$ and $X5$ are aligned to the glucose mass in plasma and in rapidly/slowly ET. Similarly, module $X13$ also presents a slightly higher $L_{INT}$ median. This module is aligned to the subcutaneous glucose level. Therefore, the model has shown a lower capacity to abstract the mechanisms for $X4$, $X5$ and $X13$ than the rest of the modules. Similar trends are observed across the other PHs, with detailed plots included in Appendix~\ref{sec:causal-abst} Table~\ref{results-iit-simplified-loss-int}.

\subsection{Sensitivity analysis with regular \textit{simglucose}}
\begin{figure}[h]
    \centering
    \includegraphics[scale=0.15]{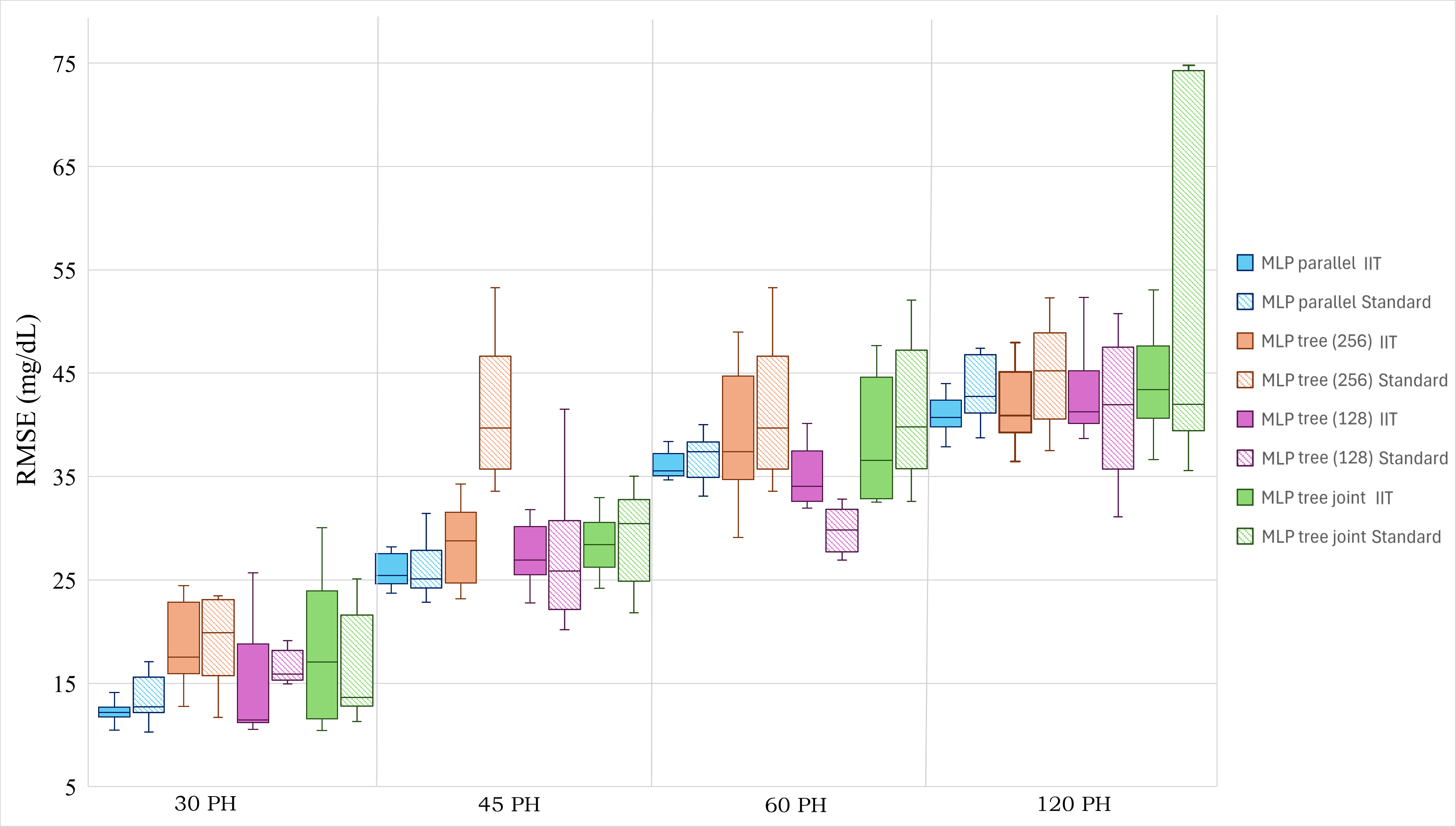}
    \caption{RMSE (mg/dL) prediction error of the different MLP architectures (4 models) across the four PHs for the test (n=30) in-silico T1DM patients. The causal model used is the regular \textit{simglucose} and the boxplots are produced with 10 different random seed. The solid bars refer to the models being trained through interchange intervention training while the striped bars refer to conventional training; without IIT. }
    \label{fig:results_iit_vs_regular}
\end{figure}

Figure~\ref{fig:results_iit_vs_regular} contrasts the four different models: MLP parallel, MLP tree (hidden sizes 128 and 256), and MLP joint; trained with IIT using regular (i.e. without removal of red terms in Figure \ref{fig:simglucose}) \textit{simglucose} as causal model and standard training. 
Detailed values for MSE, MAE and EGA performance are reported in Appendix~\ref{table:acyclic}.
For Clarke error grid analysis, the model MLP parallel Figs.~\ref{subfig:parallel-iit},~\ref{subfig:parallel-standard} show a tendency to predict within the range of 140 and 180, nearly horizontal. On the other hand, the rest of the models show more dispersed predictions along the diagonal. This holds for both IIT-trained and standard-trained models.

Overall, no model outperforms others consistently across all PHs or metrics (lowest MSE, MAE, RMSE, or highest EGA A-B). In addition, none of the four models presents a better performance defined as improvement in all metrics over the four PHs when IIT is applied to the regular \textit{simglucose} with respect to its standard training. Generally, errors increase as PH increases.

\subsection{Comparison with previous studies}
We performed a comparative analysis with prior research on BG for T1DM employing prediction models that incorporate expert knowledge. The findings are reported in Table~\ref{results-comparison}. For this comparison, we consider the MSE as the only metric due to the unavailability of MAE and the percentage of predictions within zones A or B of the EGA in most previous studies. The best results per PH are emphasized in bold.
The table shows in the first row the results obtained from the MLP tree model (with 256 as hidden size)  trained through IIT and using the regular \textit{simglucose} as causal model. Although this model is not the best-performing model in this study, it is the best model using the regular \textit{simglucose} and therefore comparable with the published models. Additionally, Appendix~\ref{ap:comparsion-studies} Table~\ref{results-comparison-granular} shows a more detailed version of Table~\ref{results-comparison}, reporting the results of each individual study that \citet{metanylis} aggregates in theirs.

\begin{table}[h]
\centering
\caption{Comparison between the model that achieved the best performance during this research MLP tree (256) IIT using regular \textit{simglucose} causal model (first row) and previous models sourced from the literature. All four prediction horizons are compared with the best results in bold.}
\label{results-comparison}
\begin{tabular}{ccccccc}
\hline
\multirow{3}{*}{Reference} & \multirow{3}{*}{\begin{tabular}[c]{@{}c@{}}Includes expert \\ knowledge \end{tabular}} & \multirow{3}{*}{\begin{tabular}[c]{@{}c@{}}UvA/Padova \\ dataset\end{tabular}} & \multicolumn{4}{c}{\begin{tabular}[c]{@{}c@{}}MSE\\ $({mg/dL})^{2}$\end{tabular}} \\ \cline{4-7} 
 &  &  & \multicolumn{1}{c|}{30 PH} & \multicolumn{1}{c|}{45 PH} & \multicolumn{1}{c|}{60 PH} & 120 PH  \\ \hline
Our model & \multicolumn{1}{c|}{Yes} & \multicolumn{1}{c|}{Yes} & \multicolumn{1}{c|}{181.50} & \multicolumn{1}{c|}{857.40} & \multicolumn{1}{c|}{1340.30} & 1521.00 \\
\citet{Lui_EGA_benchmark} & \multicolumn{1}{c|}{Yes} & \multicolumn{1}{c|}{Yes} & \multicolumn{1}{c|}{\textbf{101.20}} & \multicolumn{1}{c|}{-} & \multicolumn{1}{c|}{508.95} & 1206.86 \\
\citet{Rebaz_EGA_benchmark} & \multicolumn{1}{c|}{Yes} & \multicolumn{1}{c|}{No} & \multicolumn{1}{c|}{-} & \multicolumn{1}{c|}{-} & \multicolumn{1}{c|}{\textbf{495.96}} & \textbf{1081.28} \\
\citet{metanylis} & \multicolumn{1}{c|}{No} & \multicolumn{1}{c|}{No} & \multicolumn{1}{c|}{457.96} & \multicolumn{1}{c|}{\textbf{452.41}} & \multicolumn{1}{c|}{900.60} & - \\
\end{tabular}
\end{table}

\section{Discussion}
The objective of the experiments is to determine if IIT can be applied in real healthcare use cases that are more complex and present different challenges than those addressed in recent publications. Since the evolution of BG levels in T1DM patients can be modelled via the \textit{simglucose} simulator, we elected to employ the latter as a causal model for the IIT training of a neural network.

Firstly, we adapted the \textit{simglucose} model by compressing the time dimension and imposed acyclicity as described in Section \ref{sec:simglucose}, obtaining a SCM. 
Figure~\ref{fig:results_iit_modified_vs_regular} shows significant improvements in IIT performance across all PHs when using this amended \textit{simglucose}. IIT achieves low errors and performs better than standard training. It is worth noting that standard training is generally more unstable than IIT, as it almost doubles the standard deviation for PHs 30, 60, and 120.
However, the overlapping boxes for each PH, especially in the interquartile range, suggest that the groups have similar medians and variability. Due to the small sample size, it was not possible to test whether the performances with and without IIT were significantly different via statistical hypothesis testing.
Regarding this results,  the absolute IIT RMSE gains may appear modest. The causal model used in this study is a simplified version of the original \textit{simglucose} model, reducing the number of input parameters per patient from 64 to 13. While this abstraction lowers computational complexity and aligns with the IIT framework, it may also introduce inaccuracies by omitting important physiological dynamics. The choice of a simplified architecture was intentional to focus on evaluating the applicability of IIT. In addition, the improvements from IIT diminish as the prediction horizon increases (e.g., at 120 minutes) due to the growing influence of unmodeled factors such as behavioral variability, physical activity, and stress, which are generally challenging to capture. 

As for the causal abstraction analysis, we observe in Figure~\ref{fig:results_loss_int_fig_short} that the $L_{INT}$ lowers as the training progresses.
This metric quantifies how much the predictive model is able to align with causal knowledge.
These graphs indicate that the MLP tree (256) model is learning from the counterfactual amended \textit{simglucose} behaviour, producing better counterfactual predictions, and therefore lowering the $L_{INT}$ after each training epoch. While the $L_{INT}$ does not reach zero, and thus the model does not achieve a complete causal abstraction, the value of $L_{INT}$ does reach single digit across all horizons, suggesting a good degree of causal abstraction. Appendix~\ref{sec:causal-abst} Table~\ref{results-iit-simplified-loss-int} reports $L_{INT}$ for each IIT MLP tree (256) module for the test dataset. 
Interestingly, we observe that for all the PHs, module $X4$ and module $X5$ tend to have a slightly higher $L_{INT}$ median than the rest of the modules. Modules $X4$ and $X5$ are aligned to the variables encoding glucose mass in plasma and in rapidly/slowly equilibrating tissues. The causal mechanism is especially complex at $dxdt_{4}$, involving the  $EGP_t$ endogenous glucose production at time t, $Rat$ glucose rate of appearance in plasma at time t, $U_{iit}$ insulin-independent glucose utilization at time t and $E_t$ renal excretion at time t. Likewise, the mechanism at $dxdt_{5}$  utilizes $U_{idt}$ insulin-dependent glucose utilization at time t. The fact that $L_{INT}$ is higher in these modules may be due to the fact that they are concerned with time-dependent factors that are endogenous to the SCM, which could be particularly influential and difficult to abstract. Similarly, module $X13$ also presents a slightly higher $L_{INT}$ median. This module is aligned to the variable representing subcutaneous glucose level, which is directly receiving input from module $X4$, thus higher $L_{INT}$ values are likely due to propagation from $X4$.

Comparing the regular and amended causal models, Appendix~\ref{fig:ega} Fig.~\ref{subfig:tree-256-sim-iit} and Fig.~\ref{subfig:tree-256-sim-standard} show differences between BG targets generated by the amended simulator and those from the regular one. These figures reveal that the absence of cyclical relationships, which are essential for capturing the feedback loops in glucose-insulin dynamics, significantly alters the model’s output, leading to targets that do not fully reflect clinical behavior in T1DM patients. Nevertheless, the deviation produced is within the safe clinical areas for the patient (EGA areas A and B). For this reason, we believe experiments with the amended model are still worth conducting and sharing.
This suggests that the amended simulator, while allowing for a successful abstraction with the MLP tree (256), may not fully reflect the glucose and insulin dynamics in T1DM patients as represented by the regular \textit{simglucose}. 

Next, we evaluated the four MLP models using the regular \textit{simglucose} as the causal model. In the case of MLP parallel model, Figure~\ref{fig:results_iit_vs_regular} shows lower prediction errors with IIT at PH 30, 60 and 120, but higher errors at PH 45 compared to standard training. This model lacks connections between modules, unlike the causal model, which may explain the similar results for both training methods. As a consequence, the model is underfitted, as seen in the horizontal prediction trends for the error grid analysis in Appendix~\ref{fig:ega} Figures~\ref{subfig:parallel-iit} and~\ref{subfig:parallel-standard}.
Regarding the MLP tree model with 256 hidden size, IIT outperforms standard training: MLP tree with IIT consistently shows lower RMSE values across all PHs compared to the standard version, just like in the case of the amended simulator (albeit with higher errors across the board). These results suggest that the connections between modules help the IIT-trained MLP tree model to make more accurate predictions. Therefore, besides the mapping, an architectural resemblance to the causal model is helpful for the training. On the other hand, the MLP tree (128) model shows slightly lower RMSE, especially at PH 30 and 60. This suggests that a larger number of nodes is needed for a module to capture the complexity of the corresponding causal mechanism.
Moving to MLP joint model, its architecture design should be addressing the cyclic connections on the regular \textit{simglucose}. However, it does not significantly outperform the MLP tree model (Table~\ref{results-iit}). These results suggest that encapsulating the cycles of the time-compressed DAG in a single module does not solve the problem.

Finally, in Table~\ref{results-comparison} we compared state-of-the-art results with our MLP tree (256) using the regular \textit{simglucose}. 
The study by \citet{Lui_EGA_benchmark} achieves superior perfomance at 30-minute PH, with a MSE difference of 80.30 $(mg/dL)^2$ in favor of their model. However, \citet{Lui_EGA_benchmark} utilizes a more constrained testing dataset with lower variability, as it includes only 10 out of the 30 available UvA/Padova in-silico patients. This limitation may impact the generalizability of their results and complicate direct performance comparisons. 45-minute PH comparison is missing as the other two studies did not report their perfomance at that PH.
At PH 45, \citep{metanylis} ranks first, aggregating results from 11 different ML models.
For 60-minute PH, \citet{Rebaz_EGA_benchmark}'s model outperforms our model but their results are based on testing only on one T1DM patient, whereas we included more patient variability, possibly explaining the performance gap.
At 120-minute PH, our performance is lower ($\triangle$MSE=439.72$(mg/dL)^2$). This result underscores the increasing complexity of longer prediction horizons, where maintaining predictive accuracy becomes more challenging.
In summary, our model does not perform as well as current state-of-the-art models, but nonetheless achieves results in the same ballpark with a very simple architecture and a more challenging test set.

Beyond predictive performance, it is important to consider the cost-benefit trade-offs of incorporating expert knowledge. First of all, the causal model we used is well-documented in the T1DM domain, reducing the effort required for model development. Secondly, the primary advantage of IIT is not merely the predictive improvement but also the alignment of the neural network with causal reasoning, enhancing interpretability and trustworthiness—critical factors in clinical applications. In addition, by integrating causal knowledge, we reduce reliance on patient-specific pharmacokinetic parameters, which are expensive and time-consuming to obtain. This trade-off could make the model more scalable and practical in real-world settings. Finally, the effort to develop a causal model is a one-time investment. Even slight improvements, when applied consistently over time on a large population, could have meaningful implications for long-term diabetes management. 

\subsection{Limitations and future work}

First, the model used as a causal model is the S2008 version, the most outdated. Using the S2008 version has provided us with a causal structure to work with but it does not faithfully reflect all real clinical scenarios. In the event that in the future S2017 becomes open to the public, it is recommended to update the causal model.
Furthermore, this project has been limited to a simple MLP architecture, to ease the exploration of different alignments and configurations. Future experiments should include the investigation of a more complex MLP architecture, such as an MLP where the time dimension is not collapsed and each time-indexed variable is modelled distinctly. This would allow for a fine-grained representation of temporal dynamics. However, we have already observed that this model introduces substantial complexity due to the increased number of variables and interdependencies, which can only be investigated in the presence of much larger datasets. It is also interesting to include in this list models designed for temporal tasks such as long short-term memory or DRNN.
As a potential extension of our current findings, one could explore the idea of deliberately injecting incorrect knowledge into the model to evaluate its robustness. This approach could provide  insights into the resilience of predictive models when faced with potential errors or inaccuracies in expert knowledge, helping to assess their reliability under less-than-ideal conditions.
We would also like to highlight that the data generated for the training comes from distributions created from 30 in-silico subjects. It is recommended to obtain the real distributions of \textit{simglucose} in the future, under a paid license. In addition, it would be appropriate to test the models on real data, such as the Ohio dataset \citep{Marling_Bunescu_2020}.
Finally, if the data allows, it is recommended to run statistical tests to determine the superiority of IIT training. 

\section{Conclusion}

This study investigated the applicability of IIT for predicting BG levels in T1DM patients using neural network models. Our primary objective was to determine whether IIT could be effectively utilized in complex healthcare scenarios, which pose greater challenges than previously explored applications. Unlike traditional simulators, a NN model capable of predicting BG levels offers the potential for computationally efficient causal abstraction, making it viable for integration into lightweight, wearable devices with insulin pumps \citep{s24196322, NAHAVANDI2022106541}. Such an approach could lead to real-time, on-device BG prediction that is both accurate and causally informed.
Although the immediate application of our methodology focuses on BG prediction, the underlying approach holds potential for broader use across healthcare domains, demonstrating how a theoretical framework like IIT can be adapted to a real-world use case with promising outcomes.

By training NN with IIT to impose the causal structure of the \textit{simglucose} model, we showed that some of the IIT-trained models obtain lower prediction errors and demonstrate a closer alignment with the causal structure, as evidenced by reduced $L_{INT}$ values. Thanks to the breakdown of the $L_{INT}$ loss into components, we were also able to recognize which aspects of the causal mechanisms were captured less well. 

In conclusion, we have provided evidence that IIT can be successfully applied to complex medical prediction tasks like BG level forecasting in T1DM patients. This work indicates a path for future research to further refine neural network architectures and alignment to causal models.

\section{Code and data availability}
The code used for this study can be accessed at: \url{https://github.com/aespogom/IIT_simglucose}. For any questions or additional information regarding the code, please contact the authors.
\acks{AE would like to thank the Medical Informatics department from UvA for the opportunity to dive into this research. The authors also thank Thomas Icard for his support and feedback on earlier drafts of this work.}

\newpage

\bibliography{my_bibliography_clear}
\newpage
\appendix

\section{Data acquisition}\label{ap:data_acquisition}

In order to generate 200 virtual subject model parameter vectors, several steps are needed. The distributions for each parameter are conditioned by age, considering children from 0 to 13 years, adolescents from 14 to 20 years, and adults older than 20 years old. Each joint distribution conforms to Gaussian probability distributions. In other words, each parameter is described by three different joint distributions; one for the age range of children, one for the age range of adolescents, and one for the age range of adults. Next, we generate a random list of ages between 0 and 100 years (child n=7, adolescent n=10, adult n=183). To sample the data, the Gaussian mixture model probability distribution from the Python \textit{sklearn} library was used for each age range. More details about UVA/Padova T1DM joint distributions can be found in \citet{uva_padova} and \citet{uva_padova_2} although Table \ref{gaussian-dist} shows the means and standard deviation of the ones included in the input dataset used for this study.
This approach is novel  but  aligns with the methodology used in the simulator, which extracts subjects from these joint distributions. It is important to note that the 200 T1DM-generated subjects were used only during the training. The original 30 T1DM simulator subjects were used for the testing, ensuring that the evaluation reflects the characteristics of the original dataset.

\begin{table}[H]
\centering
\caption{Mean and standard deviation of each joint distribution for the parameters included in the data input. Each parameter is described by three different distributions depending on the age range.}
\label{gaussian-dist}
\begin{tabular}{ccccccc}
\hline
 & \multicolumn{2}{c}{Child} & \multicolumn{2}{c}{Adolescent} & \multicolumn{2}{c}{Adult} \\ \hline
\multicolumn{1}{c|}{} & Mean & \multicolumn{1}{c|}{SD} & Mean & \multicolumn{1}{c|}{SD} & Mean & SD \\ \hline
\multicolumn{1}{c|}{$x_{4}$} & 260.42 & \multicolumn{1}{c|}{25.16} & 282.12 & \multicolumn{1}{c|}{26.33} & 258.58 & 12.90 \\
\multicolumn{1}{c|}{$x_{5}$} & 95.66 & \multicolumn{1}{c|}{35.97} & 371.76 & \multicolumn{1}{c|}{17.92} & 194.25 & 29.02 \\
\multicolumn{1}{c|}{$x_{6}$} & 5.60 & \multicolumn{1}{c|}{1.23} & 5.08 & \multicolumn{1}{c|}{1.58} & 5.94 & 1.25 \\
\multicolumn{1}{c|}{$x_{8}$} & 106.72 & \multicolumn{1}{c|}{11.31} & 109.18 & \multicolumn{1}{c|}{8.82} & 105.03 & 16.71 \\
\multicolumn{1}{c|}{$x_{9}$} & 106.72 & \multicolumn{1}{c|}{11.31} & 109.18 & \multicolumn{1}{c|}{8.82} & 105.03 & 16.71 \\
\multicolumn{1}{c|}{$x_{10}$} & 2.92 & \multicolumn{1}{c|}{1.30} & 3.44 & \multicolumn{1}{c|}{1.10} & 3.50 & 1.19 \\
\multicolumn{1}{c|}{$x_{11}$} & 67.33 & \multicolumn{1}{c|}{15.78} & 67.87 & \multicolumn{1}{c|}{9.30} & 87.84 & 33.76 \\
\multicolumn{1}{c|}{$x_{12}$} & 60.43 & \multicolumn{1}{c|}{22.37} & 66.62 & \multicolumn{1}{c|}{24.62} & 112.09 & 30.73 \\
\multicolumn{1}{c|}{$x_{13}$} & 260.42 & \multicolumn{1}{c|}{25.16} & 282.11 & \multicolumn{1}{c|}{26.33} & 258.58 & 12.90 \\ \hline
\end{tabular}
\end{table}

\section{Training configuration and evaluation metrics}\label{ap:experimental_setup}

The settings used to obtain the results of the experiments are now described. The implementation of a PyTorch Trainer class orchestrates the training process for a NN given a causal model. It includes functionality for performing forward passes, optimizing the model parameters, and evaluating the trained model.

Regarding the optimizer, AdamW optimizer is used with a learning rate of 0.01, epsilon of $1e^{-6}$, and betas parameters (0.9, 0.98). The learning rate scheduler is based on linear scheduling with warm-up steps. The optimizer is minimizing the Mean Squared Error (MSE) loss. In the case of IIT experiments, causal loss and standard loss have a 0.75 and a 0.25 coefficient, respectively. In the case of standard trainings, the coefficients are 0 and 1, respectively. These coefficients represent how much each loss contributes to the global model loss. The higher the coefficient, the more effort would be dedicated to minimizing the corresponding loss. In short, a conventional training would have a causal loss coefficient of 0 and a standard loss coefficient of 1 while a purely IIT would have a causal loss coefficient of 1 and a standard loss coefficient of 0. Also, an early stopping functionality is incorporated to halt training if the validation loss does not improve for 20 consecutive epochs. The maximum number of epochs is 300.
In terms of batch size, it is set to 2 so that the interchange intervention is performed between two input patients. However, the gradient accumulation steps value is 20 so the loss is computed as if the batch size was 20. Also, gradient clipping is applied to prevent exploding gradients during backpropagation. 
A set of 10 random seeds were used to calculate the mean of the estimate and the standard deviation. The selection of these seed values is random and ensures a different initialization for each experiment.

The following performance metrics were used in the evaluation.

    - Average absolute error (MAE) is defined as the magnitude of difference between the prediction of an observation $y_i$ and the true value of that observation $\hat{y}_i$ within a dataset of length $n$.

\begin{equation} \label{MAE}
\text{MAE} = \frac{1}{n} \sum_{i=1}^{n} |y_i - \hat{y}_i|
\end{equation}

    - MSE measures the average squared difference between the predicted $y_i$ and the actual target values $\hat{y}_i$ within a dataset of length $n$.

\begin{equation} \label{MSE}
\text{MSE} = \frac{1}{n} \sum_{i=1}^{n} (y_i - \hat{y}_i)^2
\end{equation}

    - RMSE measures the square root of the squared difference between predicted values and the actual target values. Essentially, RMSE is a measure of the magnitude of the residuals, making interpretation straightforward.

\begin{equation} \label{RMSE}
RMSE = \sqrt{MSE}
\end{equation}
    
    - Percentage of predicted values in the ‘clinically acceptable’ EGA classes A and B \citep{ega-paper}. CG-EGA classifies predictions into 5 classes A-B-C-D-E with respect to the clinical outcome based on the predicted BG level. The error grid divides the plot into five regions. Points in Zone A are within 20\% of the reference sensor, showing a clinically accurate zone. Points in Zone B would not lead to inappropriate treatment. Points in Zone C would lead to unnecessary treatment, whereas the points in Zone D are potentially dangerous, and failed to detect hypoglycemia or hyperglycemia events correctly. A predicted value is termed \enquote{clinically acceptable} if it is classified into either the A or B EGA class. CG-EGA is a variation of the EGA grid in which the \enquote{accurate} domain is roughly equivalent to the EGA \enquote{clinically acceptable} classification \citet{Clarke_EGA}.

The predictions were evaluated on the 30, 45, 60 and 120-minute horizon.
The EGA and RMSE results were evaluated for completeness and to comply with GLYFE benchmark \citet{benchmark_GLYFE_2021}.

\newpage
\section{Performance using with amended \textit{simglucose} causal model}\label{table:acyclic}

Table~\ref{results-iit-simplified} presents the results for the MLP tree model with 256 hidden units, using amended \textit{simglucose} as causal model. The best results for each PH are highlighted, comparing IIT training with standard training. Mean and standard deviation are reported due to different random seeds used.
\begin{table}[H]
\caption{Results of the MLP tree (256) model across the four PHs for the test (n=30) in-silico T1DM patients (best results in bold). 10 different seeds were used to obtain the estimations. The causal model used is the amended \textit{simglucose}. The \enquote{IIT} column refers to the models being trained through interchange intervention training while \enquote{Standard} refers to conventional training; without IIT, and $\triangle$ to the difference of the means between the IIT and Standard result. A green $\triangle$ indicates that IIT achieves better performance than standard training for that metric. A red $\triangle$ indicates that IIT achieves worse performance than standard training for that metric.}
\label{results-iit-simplified}
\resizebox{\textwidth}{!}{%
    \begin{tabular}{cccccccccccccccc}
    \hline
    & \multicolumn{5}{c}{MSE $({mg/dL})^{2}$} & \multicolumn{5}{c}{MAE $({mg/dL})$} & \multicolumn{5}{c}{ EGA A-B $(\%)$} \\ \hline
    \multicolumn{1}{c|}{\multirow{2}{*}{PH}} & \multicolumn{2}{c}{IIT} & \multicolumn{2}{c}{Standard}  & \multicolumn{1}{c|}{\multirow{2}{*}{$\triangle$}} & \multicolumn{2}{c}{IIT} & \multicolumn{2}{c}{Standard}  & \multicolumn{1}{c|}{\multirow{2}{*}{$\triangle$}} & \multicolumn{2}{c}{IIT} & \multicolumn{2}{c}{Standard}  & \multirow{2}{*}{$\triangle$} \\
    
    \multicolumn{1}{c|}{} & \multicolumn{2}{c}{(Mean $\pm$ ST)} & \multicolumn{2}{c}{(Mean $\pm$ ST)} & \multicolumn{1}{c|}{} & \multicolumn{2}{c}{(Mean $\pm$ ST)} & \multicolumn{2}{c}{(Mean $\pm$ ST)} & \multicolumn{1}{c|}{} & \multicolumn{2}{c}{(Mean $\pm$ ST)} & \multicolumn{2}{c}{(Mean $\pm$ ST)} & \\ \hline

     \multicolumn{1}{c|}{30} & \textbf{255.60} & \textbf{63.83} & 301.47 & 118.15 & \multicolumn{1}{c|}{\cellcolor{green!25}{-45.87}} & \textbf{12.04} & \textbf{1.67} & 12.93 & 2.87 & \multicolumn{1}{c|}{\cellcolor{green!25}{-0.89}} & \textbf{99.67} & \textbf{1.00} & 99.00 & 1.53 & \cellcolor{green!25}{0.67} \\ \hline
    
    \multicolumn{1}{c|}{45} & \textbf{538.63} & \textbf{133.18} & 560.33 &  126.24  & \multicolumn{1}{c|}{\cellcolor{green!25}{-21.69}} & \textbf{16.98} & \textbf{2.31} & 17.55 & 1.82 & \multicolumn{1}{c|}{\cellcolor{green!25}{-0.58}} & \textbf{99.33} & \textbf{2.00} & 98.33 & 2.24  & \cellcolor{green!25}{1.00} \\ \hline
    
     \multicolumn{1}{c|}{60} & \textbf{885.20} & \textbf{175.13} & 981.09 & 327.36 & \multicolumn{1}{c|}{\cellcolor{green!25}{-95.89}}  & \textbf{22.01} & \textbf{2.85} & 23.30 & 5.12 & \multicolumn{1}{c|}{\cellcolor{green!25}{-1.29}} & 99.67 & 1.00 & 99.67 & 1.00  & 0 \\ \hline
    
     \multicolumn{1}{c|}{120} & \textbf{642.95} & \textbf{85.93} & 704.50 & 202.76 & \multicolumn{1}{c|}{\cellcolor{green!25}{-128.34}} & \textbf{19.03} & \textbf{1.57} & 19.39 & 3.50 & \multicolumn{1}{c|}{\cellcolor{green!25}{-0.36}} & \textbf{99.33} & \textbf{2.00} & 98.67 & 2.21 & \cellcolor{green!25}{0.66} \\ \hline
    \end{tabular}
}
\end{table}

\section{Comparison with previous studies}\label{ap:comparsion-studies}
\begin{table}[H]
\centering
\caption{Granular comparison between the model that achieved the best performance during this research MLP tree (256) IIT using regular \textit{simglucose} causal model (first row) and previous models sourced from the literature. All four prediction horizons are compared with the best results in bold.}
\label{results-comparison-granular}
\begin{tabular}{ccccccc}
\hline
\multirow{3}{*}{Reference} & \multirow{3}{*}{\begin{tabular}[c]{@{}c@{}}Includes expert \\ knowledge \end{tabular}} & \multirow{3}{*}{\begin{tabular}[c]{@{}c@{}}UvA/Padova \\ dataset\end{tabular}} & \multicolumn{4}{c}{\begin{tabular}[c]{@{}c@{}}MSE\\ $({mg/dL})^{2}$\end{tabular}} \\ \cline{4-7} 
 &  &  & \multicolumn{1}{c|}{30 PH} & \multicolumn{1}{c|}{45 PH} & \multicolumn{1}{c|}{60 PH} & 120 PH  \\ \hline
Our model & \multicolumn{1}{c|}{Yes} & \multicolumn{1}{c|}{Yes} & \multicolumn{1}{c|}{181.50} & \multicolumn{1}{c|}{857.40} & \multicolumn{1}{c|}{1340.30} & 1521.00 \\
\citet{Lui_EGA_benchmark} & \multicolumn{1}{c|}{Yes} & \multicolumn{1}{c|}{Yes} & \multicolumn{1}{c|}{101.20} & \multicolumn{1}{c|}{-} & \multicolumn{1}{c|}{508.95} & 1206.86 \\
\citet{Rebaz_EGA_benchmark} & \multicolumn{1}{c|}{Yes} & \multicolumn{1}{c|}{No} & \multicolumn{1}{c|}{-} & \multicolumn{1}{c|}{-} & \multicolumn{1}{c|}{\textbf{495.96}} & \textbf{1081.28} \\
\citet{perez} & \multicolumn{1}{c|}{No}  & \multicolumn{1}{c|}{No} & \multicolumn{1}{c|}{324.00} & \multicolumn{1}{c|}{729.00} & \multicolumn{1}{c|}{-} & - \\
\citet{arima} & \multicolumn{1}{c|}{No}  & \multicolumn{1}{c|}{No} & \multicolumn{1}{c|}{490.62} & \multicolumn{1}{c|}{-} & \multicolumn{1}{c|}{-} & - \\
\citet{Zhu} & \multicolumn{1}{c|}{No}  & \multicolumn{1}{c|}{No} & \multicolumn{1}{c|}{357.21} & \multicolumn{1}{c|}{-} & \multicolumn{1}{c|}{-} & - \\
\citet{DANTONI2020106134} & \multicolumn{1}{c|}{No}  & \multicolumn{1}{c|}{No} & \multicolumn{1}{c|}{349.69} & \multicolumn{1}{c|}{-} & \multicolumn{1}{c|}{-} & - \\
\citet{Amar} & \multicolumn{1}{c|}{No}  & \multicolumn{1}{c|}{No} & \multicolumn{1}{c|}{561.22} & \multicolumn{1}{c|}{-} & \multicolumn{1}{c|}{1706.52} & - \\
\citet{Li} & \multicolumn{1}{c|}{No}  & \multicolumn{1}{c|}{Yes} & \multicolumn{1}{c|}{115.13} & \multicolumn{1}{c|}{-} & \multicolumn{1}{c|}{513.02} & - \\
\citet{Zecc} & \multicolumn{1}{c|}{No}  & \multicolumn{1}{c|}{Yes} & \multicolumn{1}{c|}{\textbf{88.36}} & \multicolumn{1}{c|}{-} & \multicolumn{1}{c|}{-} & - \\
\citet{Mohebbi} & \multicolumn{1}{c|}{No}  & \multicolumn{1}{c|}{No} & \multicolumn{1}{c|}{433.61} & \multicolumn{1}{c|}{851.06} & \multicolumn{1}{c|}{1340.09} & - \\
\citet{Daniels2022} & \multicolumn{1}{c|}{No}  & \multicolumn{1}{c|}{No} & \multicolumn{1}{c|}{353.44} & \multicolumn{1}{c|}{\textbf{640.09}} & \multicolumn{1}{c|}{1011.24} & 2227.84 \\
\end{tabular}
\end{table}

\newpage
\section{Causal abstraction analysis}\label{sec:causal-abst}
\begin{figure}[H]
    \centering
    \subfigure[]{\includegraphics[width=0.49\textwidth]{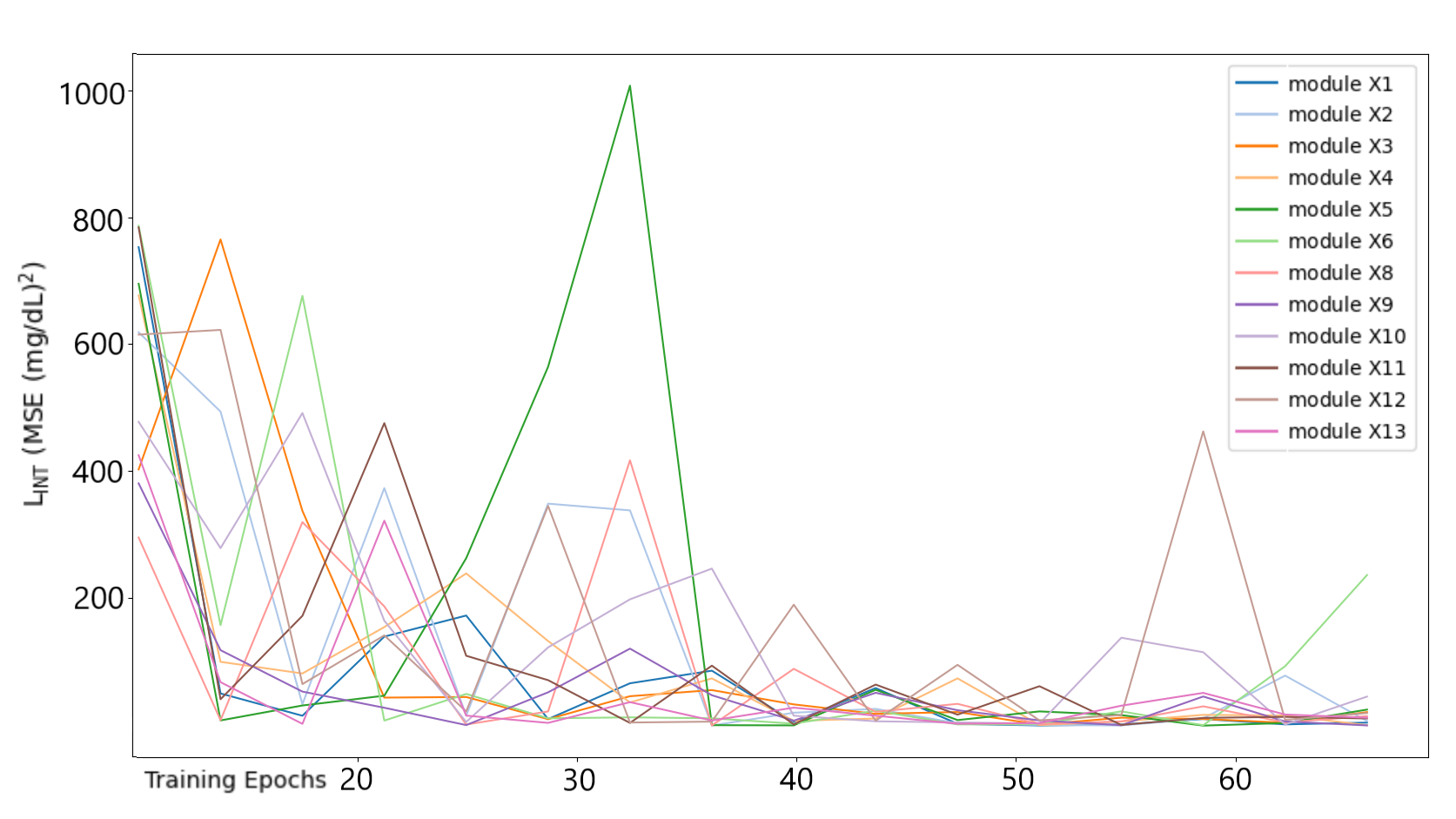}} 
    \subfigure[]{\includegraphics[width=0.49\textwidth]{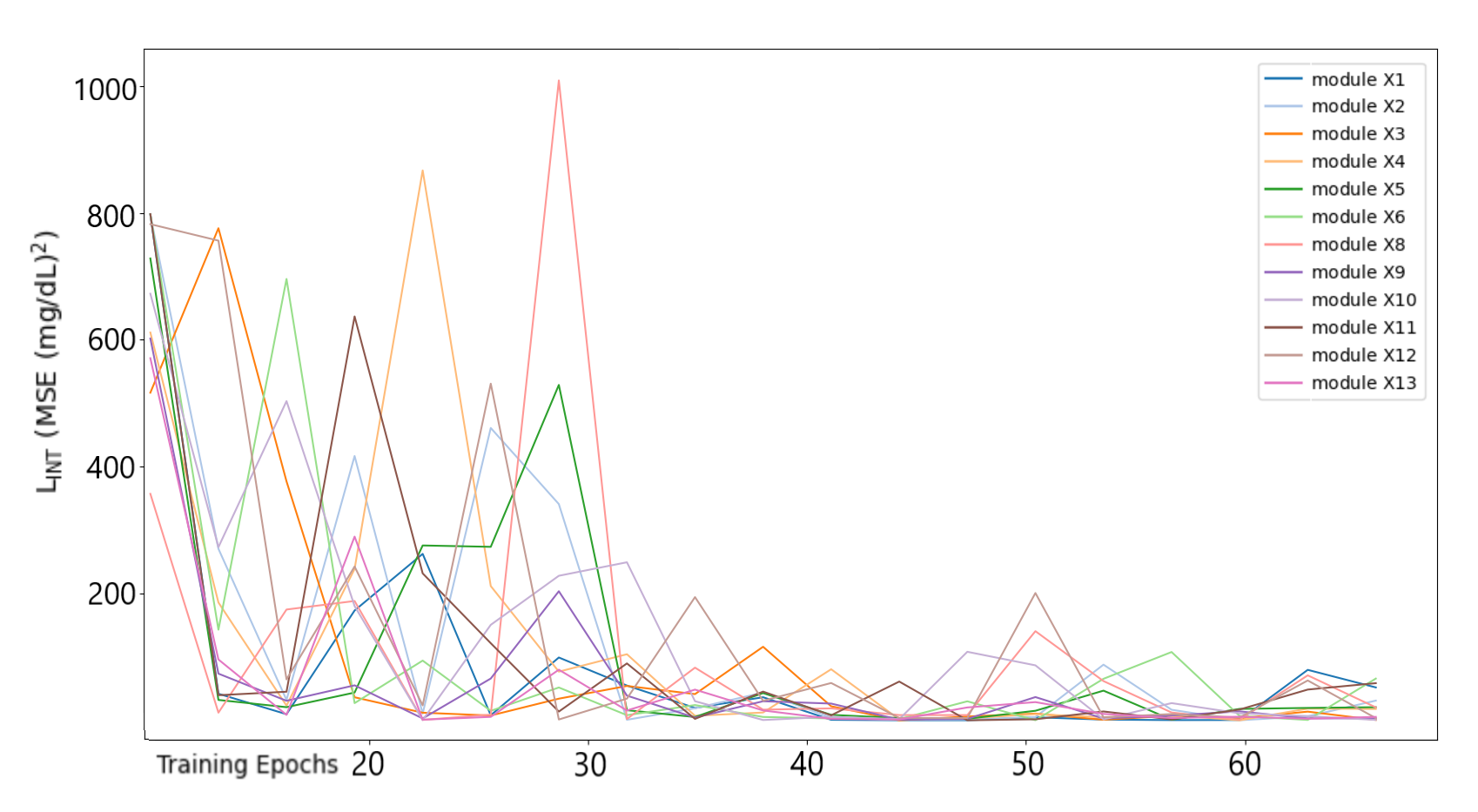}} 
    \hfill
    \subfigure[]{\includegraphics[width=0.49\textwidth]{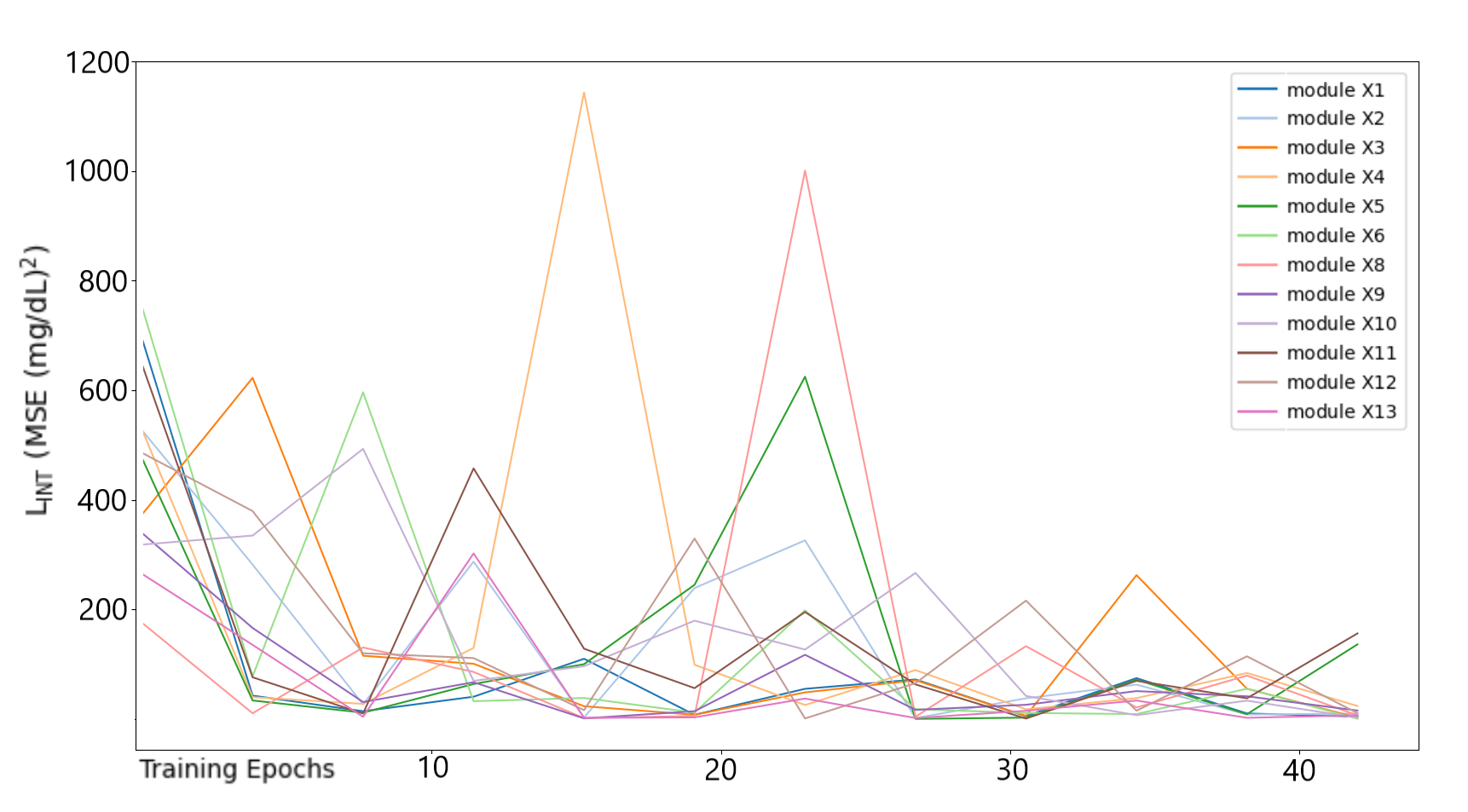}}
    \subfigure[]{\includegraphics[width=0.49\textwidth]{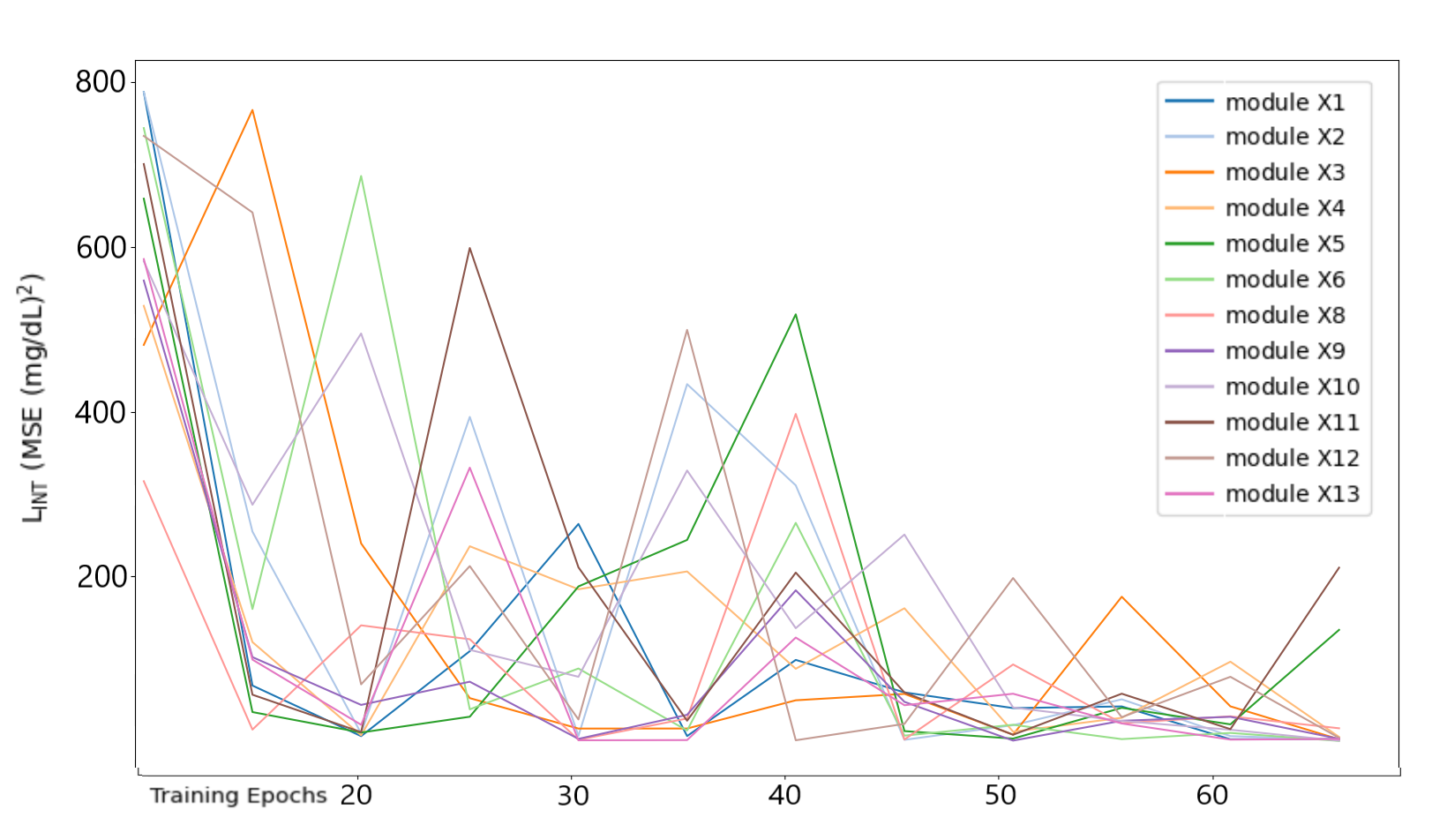}}
    \caption{MLP IIT $L_{INT} (MSE(mg/dL)2)$ during the training using the amended \textit{simglucose} as the causal model for PHs (a) 30, (b) 45, (c) 60 and (d) 120. The $L_{INT}$ is grouped by modules.}
  \label{fig:results_loss_int_fig}
\end{figure}

\begin{figure}[H]
    \centering
    \subfigure[]{\includegraphics[width=0.49\textwidth]{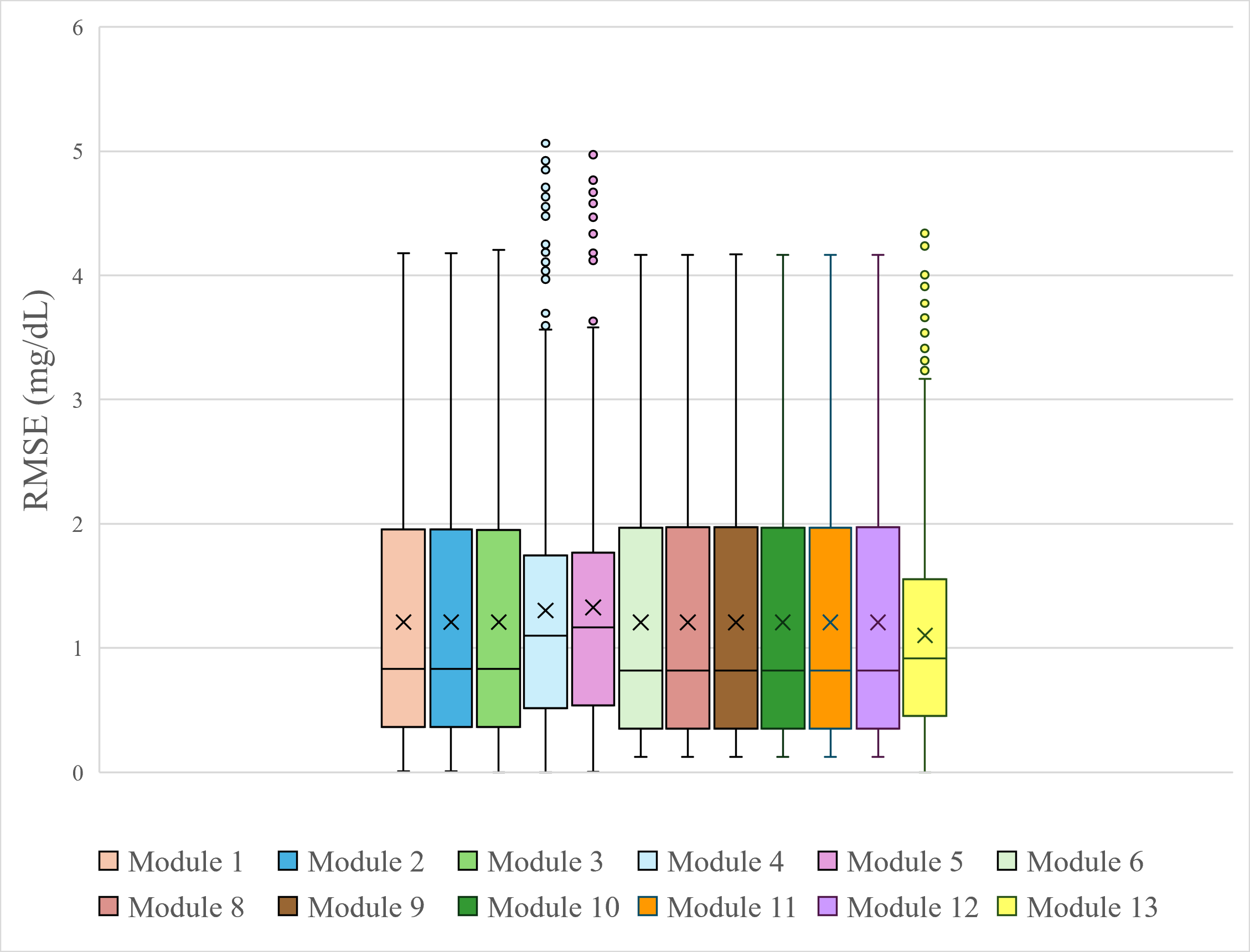}} 
    \subfigure[]{\includegraphics[width=0.49\textwidth]{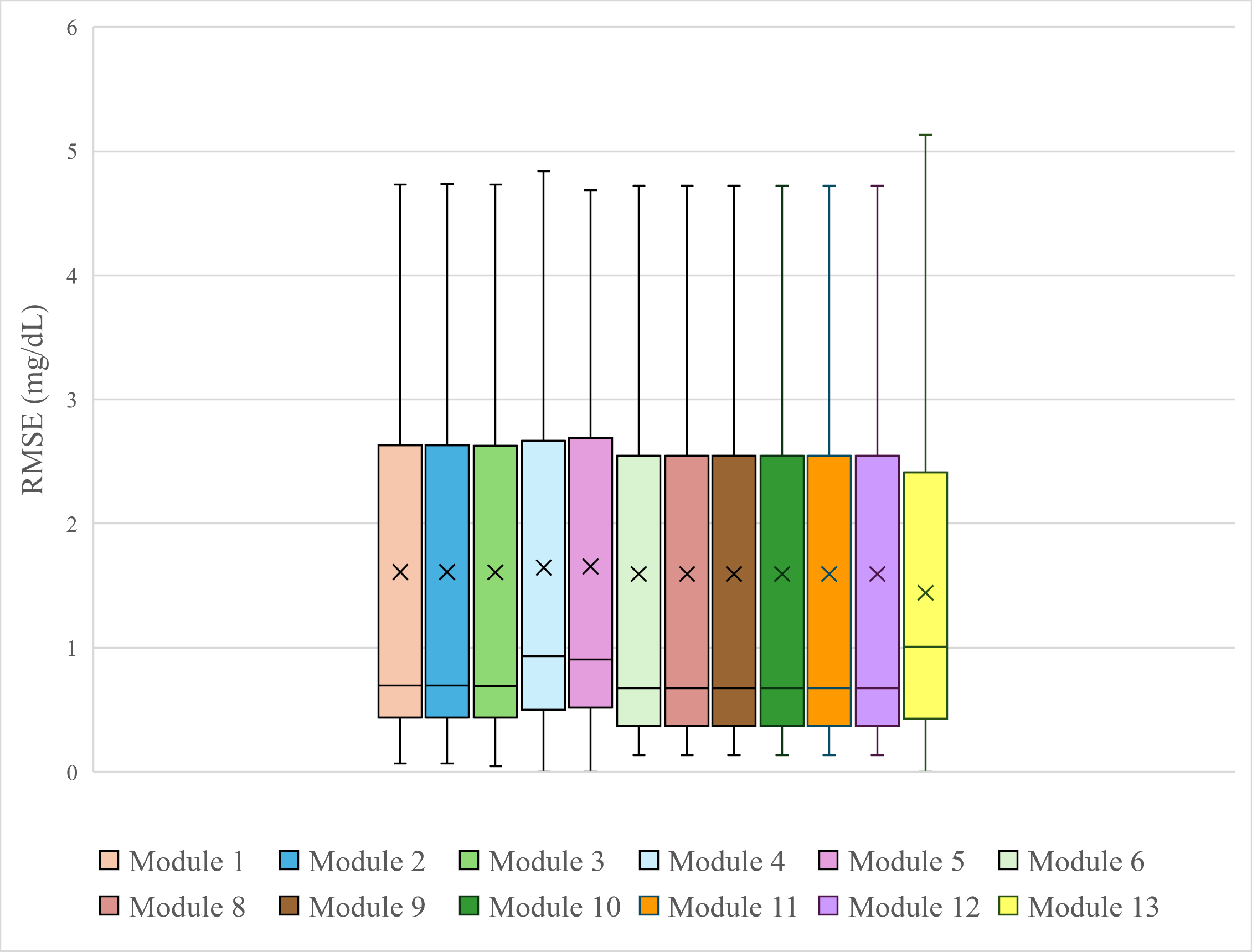}} 
    \hfill
    \subfigure[]{\includegraphics[width=0.49\textwidth]{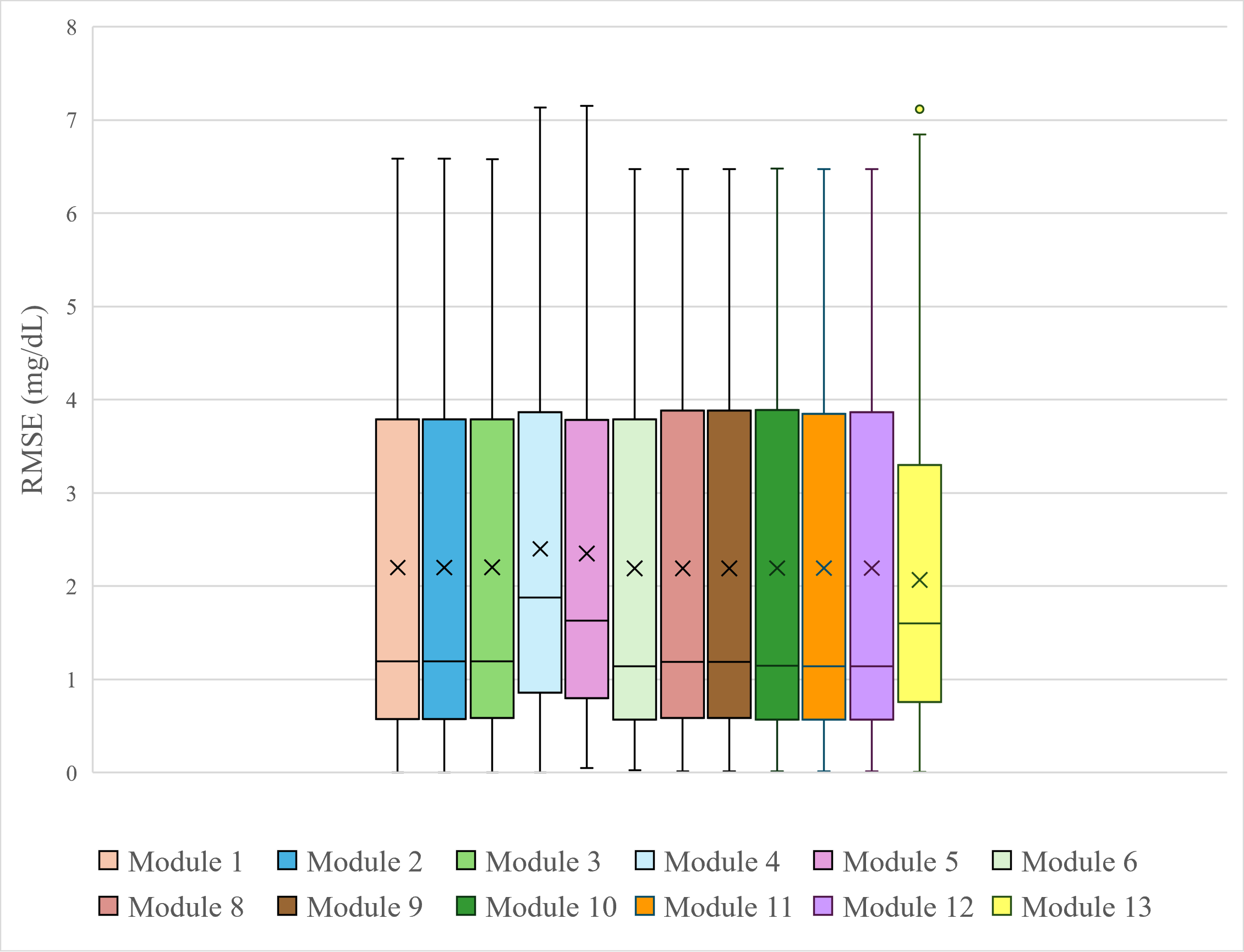}}
    \subfigure[]{\includegraphics[width=0.49\textwidth]{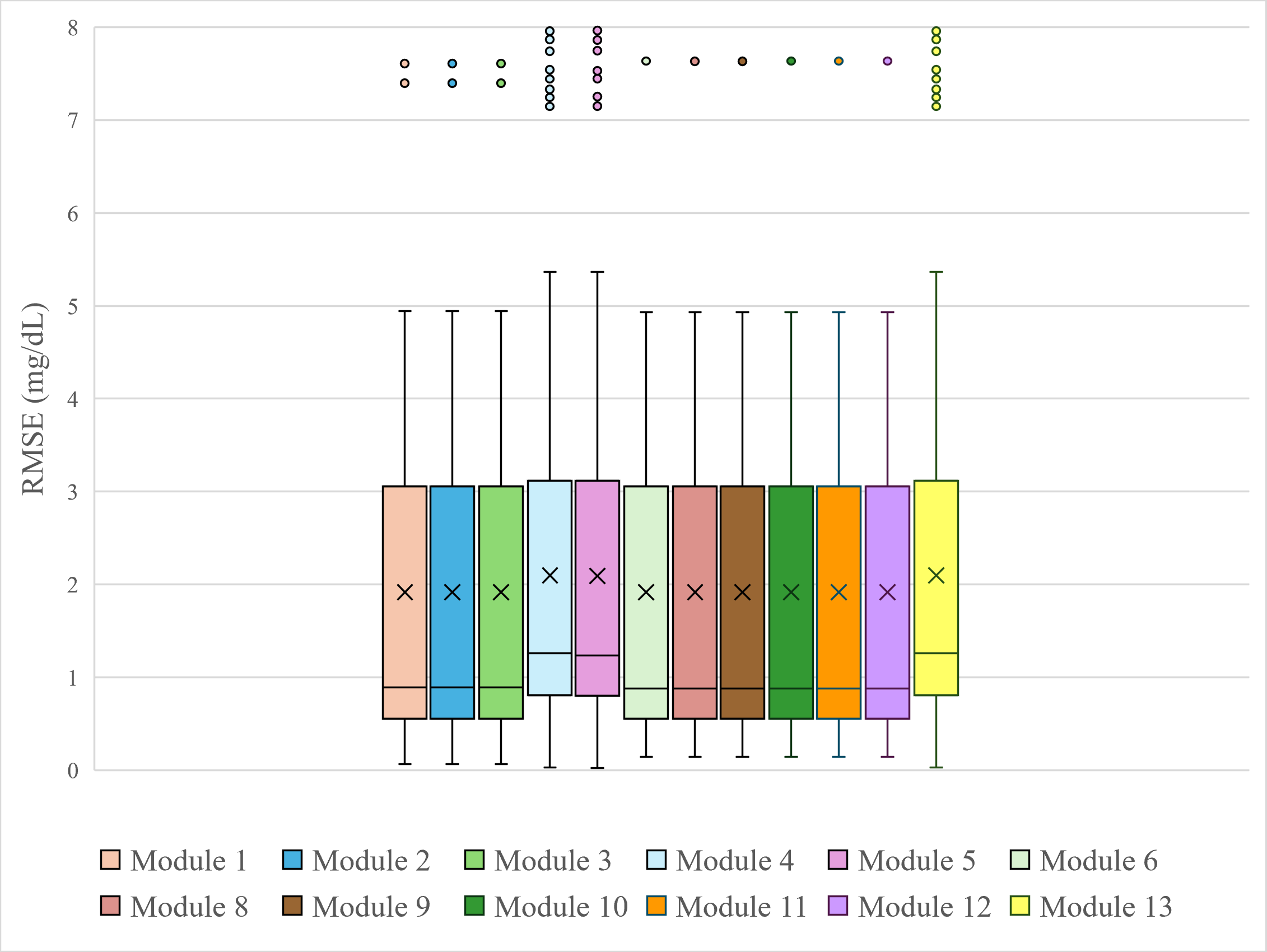}}
    \caption{MLP IIT $L_{INT}$ during the testing using the amended \textit{simglucose} as the causal model for PHs (a) 30, (b) 45, (c) 60 and (d) 120. The $L_{INT}$ is grouped by modules.}
    \label{fig:results_loss_test_int_figs}
\end{figure}

Counterfactual loss ($L_{INT}$) during the training (Fig.~\ref{fig:results_loss_int_fig}) and testing (Table~\ref{results-iit-simplified-loss-int} and Fig.~\ref{fig:results_loss_test_int_figs}) processes across different PHs for the MLP tree model with 256 hidden units, using the amended \textit{simglucose}. This involved interchanging interventions between the causal model and MLP tree model, calculating the loss between their output values.

\newpage

\begin{table}[H]
\centering
\caption{$L_{INT}$ results of the MLP tree (256) model across the four prediction horizons (PH) for the test (n=30) in-silico T1DM patients. The random seed is 56. The causal model used is the DAG \textit{simglucose}. All interchanged interventions are performed for the models trained through IIT, and then the loss for the two outputs is calculated. The lower $L_{INT}$, the higher the causal abstraction is achieved.}
\label{results-iit-simplified-loss-int}
\begin{tabular}{ccccccccc}
\hline
\multicolumn{9}{c}{$L_{INT}$ (RMSE $mg/dL$)} \\ \hline
PH & \multicolumn{2}{c}{30} & \multicolumn{2}{c}{45} & \multicolumn{2}{c}{60} & \multicolumn{2}{c}{120} \\ \hline
\multicolumn{1}{c|}{Module} & \multicolumn{2}{c|}{Mean $\pm$ SD} & \multicolumn{2}{c|}{Mean $\pm$ SD} & \multicolumn{2}{c|}{Mean $\pm$ SD} & \multicolumn{2}{c}{Mean $\pm$ SD} \\ \cline{2-9} 

\multicolumn{1}{c|}{X1} & \multicolumn{2}{c|}{1.21 $\pm$ 1.04} & \multicolumn{2}{c|}{1.61 $\pm$ 1.52} & \multicolumn{2}{c|}{2.20 $\pm$ 1.95} & \multicolumn{2}{c}{1.92 $\pm$ 1.75} \\

\multicolumn{1}{c|}{X2} & \multicolumn{2}{c|}{1.21 $\pm$ 1.04} & \multicolumn{2}{c|}{1.61 $\pm$ 1.52} & \multicolumn{2}{c|}{2.20 $\pm$ 1.95} & \multicolumn{2}{c}{1.92 $\pm$ 1.75} \\

\multicolumn{1}{c|}{X3} & \multicolumn{2}{c|}{1.21 $\pm$ 1.04} & \multicolumn{2}{c|}{1.64 $\pm$ 1.43} & \multicolumn{2}{c|}{ 2.20 $\pm$ 1.95} & \multicolumn{2}{c}{1.92 $\pm$ 1.75}\\

\multicolumn{1}{c|}{X4} & \multicolumn{2}{c|}{1.30 $\pm$ 1.01} & \multicolumn{2}{c|}{1.65 $\pm$ 1.47} &\multicolumn{2}{c|}{2.40 $\pm$ 1.84} & \multicolumn{2}{c}{2.10 $\pm$ 1.70} \\

\multicolumn{1}{c|}{X5} & \multicolumn{2}{c|}{1.33 $\pm$ 1.02} & \multicolumn{2}{c|}{1.60 $\pm$ 1.53} & \multicolumn{2}{c|}{2.35 $\pm$ 1.87} & \multicolumn{2}{c}{2.09 $\pm$ 1.70} \\

\multicolumn{1}{c|}{X6} & \multicolumn{2}{c|}{1.21 $\pm$ 1.05} & \multicolumn{2}{c|}{1.60 $\pm$ 1.53} & \multicolumn{2}{c|}{2.19 $\pm$ 1.99} & \multicolumn{2}{c}{1.92 $\pm$ 1.78} \\

\multicolumn{1}{c|}{X8} & \multicolumn{2}{c|}{1.21 $\pm$ 1.05} & \multicolumn{2}{c|}{1.60 $\pm$ 1.53} & \multicolumn{2}{c|}{2.19 $\pm$ 1.95} & \multicolumn{2}{c}{1.92 $\pm$ 1.78} \\

\multicolumn{1}{c|}{X9} & \multicolumn{2}{c|}{1.21 $\pm$ 1.05} & \multicolumn{2}{c|}{1.60 $\pm$ 1.53} & \multicolumn{2}{c|}{2.19 $\pm$ 1.95} & \multicolumn{2}{c}{1.92 $\pm$ 1.78} \\

\multicolumn{1}{c|}{X10} & \multicolumn{2}{c|}{1.21 $\pm$ 1.05} & \multicolumn{2}{c|}{1.60 $\pm$ 1.53} & \multicolumn{2}{c|}{2.19 $\pm$ 1.98} & \multicolumn{2}{c}{1.92 $\pm$ 1.78} \\

\multicolumn{1}{c|}{X11} & \multicolumn{2}{c|}{1.21 $\pm$ 1.05} & \multicolumn{2}{c|}{1.60 $\pm$ 1.53} &\multicolumn{2}{c|}{2.19 $\pm$ 1.98} & \multicolumn{2}{c}{1.92 $\pm$ 1.78} \\

\multicolumn{1}{c|}{X12} & \multicolumn{2}{c|}{1.21 $\pm$ 1.05} & \multicolumn{2}{c|}{1.60 $\pm$ 1.53} & \multicolumn{2}{c|}{2.19 $\pm$ 1.98} & \multicolumn{2}{c}{1.92 $\pm$ 1.78} \\

\multicolumn{1}{c|}{X13} & \multicolumn{2}{c|}{1.10 $\pm$ 0.84} & \multicolumn{2}{c|}{1.44 $\pm$ 1.20} & \multicolumn{2}{c|}{2.06 $\pm$ 1.57} & \multicolumn{2}{c}{2.10 $\pm$ 1.70} \\ \hline
\end{tabular}
\end{table}

\newpage
\section{Performance using the regular \textit{simglucose} causal model}\label{table:cyclic}
Table~\ref{results-iit} presents the results for the MLP tree model with 256 hidden units, using regular \textit{simglucose} as causal model. The best results for each PH are highlighted, comparing IIT training with standard training.

\begin{table}[H]
\caption{Results of the four different MLP models (Parallel, Tree 256, Tree 128, Joint) models across the four PHs for the test (n=30) in-silico T1DM patients (best results in bold). 10 different seeds were used to obtain the estimations. The causal model used is the regular \textit{simglucose}. The \enquote{IIT} column refers to the models being trained through interchange intervention training while \enquote{Standard} refers to conventional training; without IIT, and $\triangle$ to the difference of means between the IIT and Standard result. A green $\triangle$ indicates that IIT achieves better performance than standard training for that metric. A red $\triangle$ indicates that IIT achieves worse performance than standard training for that metric.}
\label{results-iit}

\resizebox{\textwidth}{!}{%
    \begin{tabular}{ccccccccccccccccc}
    \hline
    & \multicolumn{6}{c}{MSE $({mg/dL})^{2}$} & \multicolumn{5}{c}{MAE $({mg/dL})$} & \multicolumn{5}{c}{ EGA A-B $(\%)$} \\ \hline
    
    \multirow{2}{*}{MLP} & \multicolumn{1}{c|}{\multirow{2}{*}{PH}} & \multicolumn{2}{c}{IIT} & \multicolumn{2}{c}{Standard}  & \multicolumn{1}{c|}{\multirow{2}{*}{$\triangle$}} & \multicolumn{2}{c}{IIT} & \multicolumn{2}{c}{Standard}  & \multicolumn{1}{c|}{\multirow{2}{*}{$\triangle$}} & \multicolumn{2}{c}{IIT} & \multicolumn{2}{c}{Standard}  & \multirow{2}{*}{$\triangle$} \\
    
    \multicolumn{2}{c|}{} & \multicolumn{2}{c}{(Mean $\pm$ ST)} & \multicolumn{2}{c}{(Mean $\pm$ ST)} & \multicolumn{1}{c|}{} & \multicolumn{2}{c}{(Mean $\pm$ ST)} & \multicolumn{2}{c}{(Mean $\pm$ ST)} & \multicolumn{1}{c|}{} & \multicolumn{2}{c}{(Mean $\pm$ ST)} & \multicolumn{2}{c}{(Mean $\pm$ ST)} & \\ \hline

    Parallel & \multicolumn{1}{c|}{30} & \textbf{149.79
} & \textbf{24.71} & 185.30 & 58.15 & \multicolumn{1}{c|}{\cellcolor{green!25}{-35.51}} & \textbf{9.83} & \textbf{1.55} & 10.25 & 1.57 & \multicolumn{1}{c|}{\cellcolor{green!25}{-8.28}} & 97.58 & 4.96 & \textbf{97.67} & \textbf{2.74} & \cellcolor{red!25}{-0.09} \\ 
    & \multicolumn{1}{c|}{45} & \textbf{666.81
    } & \textbf{80.37} & 680.52 & 140.10 & \multicolumn{1}{c|}{\cellcolor{green!25}{-13.71}} & \textbf{19.52} & \textbf{1.65} & 20.15 & 2.29 & \multicolumn{1}{c|}{\cellcolor{green!25}{-0.63}} & \textbf{79.00} & \textbf{3.53} & 76.67 & 7.37 & \cellcolor{green!25}{2.33} \\ 
    & \multicolumn{1}{c|}{60} & \textbf{1301.95
    } & \textbf{88.66} & 1359.00 & 159.30 & \multicolumn{1}{c|}{\cellcolor{green!25}{-93.05}} & \textbf{27.36} & \textbf{1.37} & 28.00 & 2.38 & \multicolumn{1}{c|}{\cellcolor{green!25}{-0.64}} & \textbf{72.33} & \textbf{2.25} & 70.33 & 3.99 & \cellcolor{green!25}{2.00} \\ 
    & \multicolumn{1}{c|}{120} & \textbf{1682.48
    } & \textbf{150.68} & 3696.86 & 5838.54 & \multicolumn{1}{c|}{\cellcolor{green!25}{-35.51}} & \textbf{31.50} & \textbf{1.64} & 43.39 & 33.00 & \multicolumn{1}{c|}{\cellcolor{green!25}{-11.89}} & \textbf{62.67} & \textbf{4.39} & 54.67 & 19.76 & \cellcolor{green!25}{8.00} \\ \hline

    Tree (256) & \multicolumn{1}{c|}{30} & \textbf{359.31
} & \textbf{144.75} & 379.01 & 145.52 & \multicolumn{1}{c|}{\cellcolor{green!25}{-19.7}} & \textbf{14.68} & \textbf{2.92} & 15.74 & 3.94 & \multicolumn{1}{c|}{\cellcolor{green!25}{-1.06}} & \textbf{90.00} & \textbf{7.70} & 89.00 & 8.02 & \cellcolor{green!25}{1.00} \\ 
    & \multicolumn{1}{c|}{45} & \textbf{814.44} & \textbf{205.59} & 1710.04 & 554.81 & \multicolumn{1}{c|}{\cellcolor{green!25}{-895.61}} & \textbf{22.87} & \textbf{3.32} & 25.57 & 5.14 & \multicolumn{1}{c|}{\cellcolor{green!25}{-2.7}} &\textbf{74.67} & \textbf{6.52} & 67.67 & 11.97 & \cellcolor{green!25}{7.00} \\ 
    & \multicolumn{1}{c|}{60} & \textbf{1563.24} & \textbf{489.79} & 1710.04 & 554.81 & \multicolumn{1}{c|}{\cellcolor{green!25}{-146.8}} & \textbf{31.17} & \textbf{6.14} & 33.10 & 6.67 & \multicolumn{1}{c|}{\cellcolor{green!25}{-1.93}} & \textbf{65.33} & \textbf{10.68} & 64.33 & 15.56 & \cellcolor{green!25}{1.00} \\ 
    & \multicolumn{1}{c|}{120} & \textbf{1756.24
    } & \textbf{302.96} & 2033.94 & 431.97 & \multicolumn{1}{c|}{\cellcolor{green!25}{-277.70}} & \textbf{32.73} & \textbf{2.95} & 35.41 & 4.47 & \multicolumn{1}{c|}{\cellcolor{green!25}{-2.68}} & \textbf{60.00} & \textbf{6.48} & 57.67 & 7.04 & \cellcolor{green!25}{2.33} \\ \hline

    Tree (128) & \multicolumn{1}{c|}{30} & \textbf{245.14
} & \textbf{179.81} & 303.68 & 124.36 & \multicolumn{1}{c|}{\cellcolor{green!25}{-58.54}} & \textbf{11.50} & \textbf{4.13} & 12.93 & 3.03 & \multicolumn{1}{c|}{\cellcolor{green!25}{-1.43}} & \textbf{95.00} & \textbf{6.14} & 82.67 & 1.41 & \cellcolor{green!25}{12.33} \\ 
    & \multicolumn{1}{c|}{45} & \textbf{905.76
    } & \textbf{528.66} & 814.23 & 492.97 & \multicolumn{1}{c|}{\cellcolor{green!25}{-91.53}} & 23.22 & 7.79 & \textbf{18.64} & \textbf{2.49} & \multicolumn{1}{c|}{\cellcolor{red!25}{4.58}} & \textbf{70.00} & \textbf{16.56} & 68.33 & 1.76 & \cellcolor{green!25}{1.67} \\ 
    & \multicolumn{1}{c|}{60} & 1284.24
     & 336.86 & \textbf{976.58} & \textbf{336.21} & \multicolumn{1}{c|}{\cellcolor{red!25}{307.66}} & 27.59 & 4.06 & \textbf{23.20} & \textbf{5.45} & \multicolumn{1}{c|}{\cellcolor{red!25}{4.39}} & \textbf{70.67} & \textbf{7.83} & 57.33 & 4.10 & \cellcolor{green!25}{13.34} \\ 
    & \multicolumn{1}{c|}{120} & 1956.33
     & 639.65 & \textbf{1771.60} & \textbf{532.03} & \multicolumn{1}{c|}{\cellcolor{red!25}{184.73}} & 34.16 & 6.47 & \textbf{19.89} & \textbf{3.62} & \multicolumn{1}{c|}{\cellcolor{red!25}{14.27}} & \textbf{58.33} & \textbf{8.50} & 58.18 & 4.05 & \cellcolor{green!25}{0.15} \\ \hline

    Joint & \multicolumn{1}{c|}{30} & 359.06
 & 255.69 & \textbf{287.95} & \textbf{176.92} & \multicolumn{1}{c|}{\cellcolor{red!25}{71.11}} & 15.00 & 6.36 & \textbf{13.46} & \textbf{4.76} & \multicolumn{1}{c|}{\cellcolor{red!25}{1.54}} & 90.33 & 15.59 & \textbf{94.67} & \textbf{8.20} & \cellcolor{red!25}{-4.34} \\ 
    & \multicolumn{1}{c|}{45} & \textbf{848.92} & \textbf{242.88} & 874.63 & 247.22 & \multicolumn{1}{c|}{\cellcolor{green!25}{-25.71}} & \textbf{23.07} & \textbf{4.12} & 23.48 & 4.43 & \multicolumn{1}{c|}{\cellcolor{green!25}{-0.41}} & 73.00 & 10.71 & 73.00 & 9.36 & 0 \\ 
    & \multicolumn{1}{c|}{60} & \textbf{1527.63
    } & \textbf{493.59} & 1718.96 & 537.99 & \multicolumn{1}{c|}{\cellcolor{green!25}{-191.33}} & \textbf{30.75} & \textbf{5.88} & 33.21 & 6.04 & \multicolumn{1}{c|}{\cellcolor{green!25}{-2.46}} & \textbf{65.67} & \textbf{10.31} & 63.33 & 7.70 & -\cellcolor{green!25}{2.34} \\ 
    & \multicolumn{1}{c|}{120} & \textbf{1979.62
    } & \textbf{462.07} & 5559.40 & 7901.60 & \multicolumn{1}{c|}{\cellcolor{green!25}{-3579.78}} & \textbf{35.39} & \textbf{4.75} & 54.01 & 44.38 & \multicolumn{1}{c|}{\cellcolor{green!25}{-18.62}} & \textbf{56.33} & \textbf{7.61} & 46.67 & 25.29 & \cellcolor{green!25}{9.66} \\ \hline

    \end{tabular}
}
\end{table}

\newpage
\section{EGAs}\label{fig:ega}
\begin{figure}[H]
    \centering
    \subfigure[]{\includegraphics[width=0.49\textwidth]{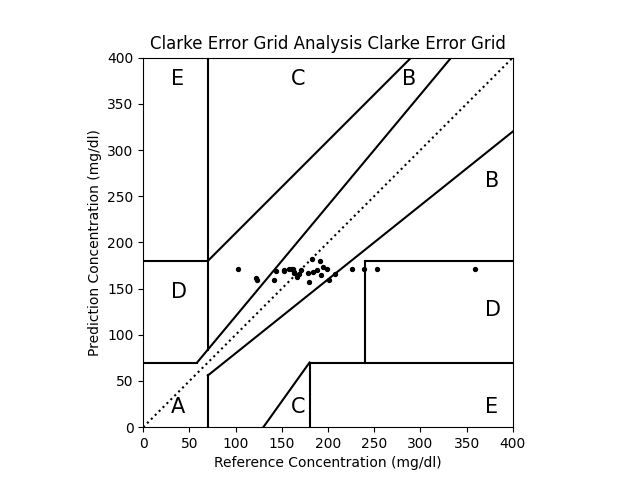}\label{subfig:parallel-iit}}
    \subfigure[]{\includegraphics[width=0.49\textwidth]{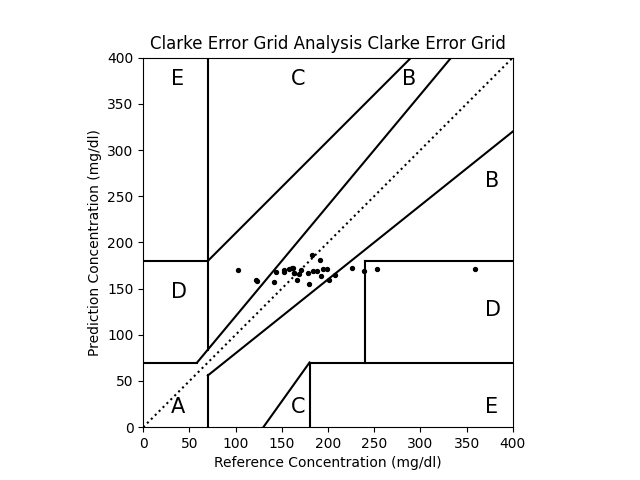}\label{subfig:parallel-standard}}
    \subfigure[]{\includegraphics[width=0.49\textwidth]{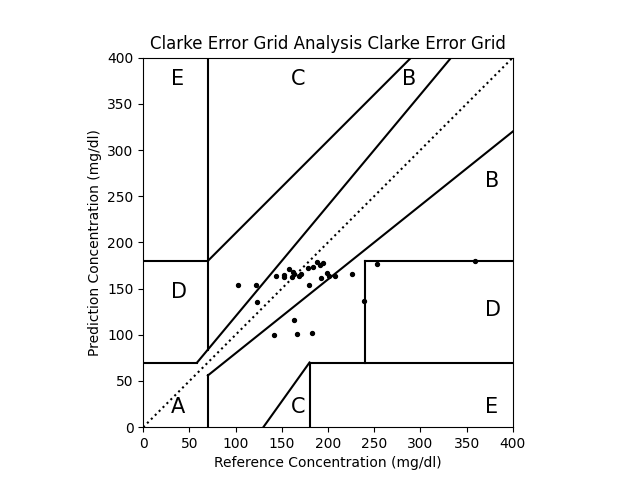}\label{subfig:joint-iit}}
    \subfigure[]{\includegraphics[width=0.49\textwidth]{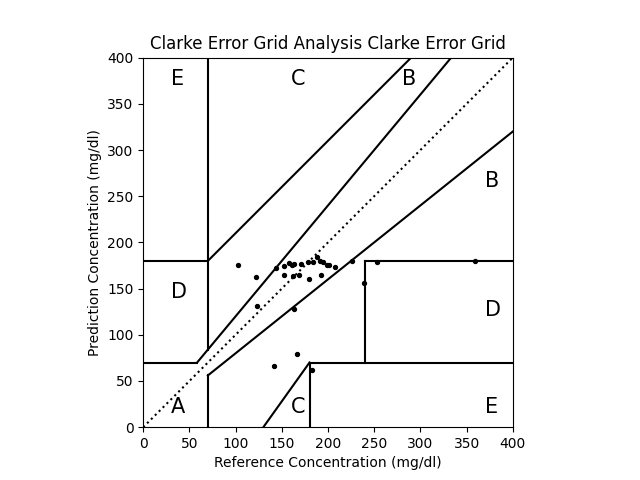}\label{subfig:joint-standard}}
    \caption{Error grid 120 PH of the MLP parallel architecture (a) and (b), and MLP joint architecture (c) and (d). The images on the left (a) and (c) refer to the models being trained through interchange intervention training using the regular causal model while the images on the right (b) and (d) correspond to conventional training; without IIT.}
\end{figure}

\begin{figure}[H]
    \centering
    \subfigure[]{\includegraphics[width=0.49\textwidth]{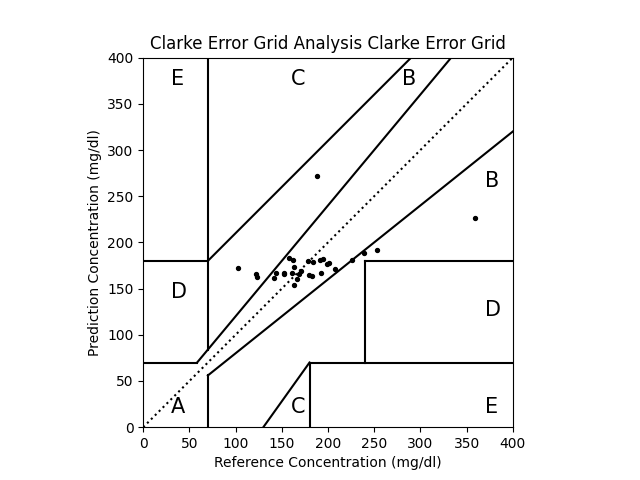}\label{subfig:tree-128-iit}}
    \subfigure[]{\includegraphics[width=0.49\textwidth]{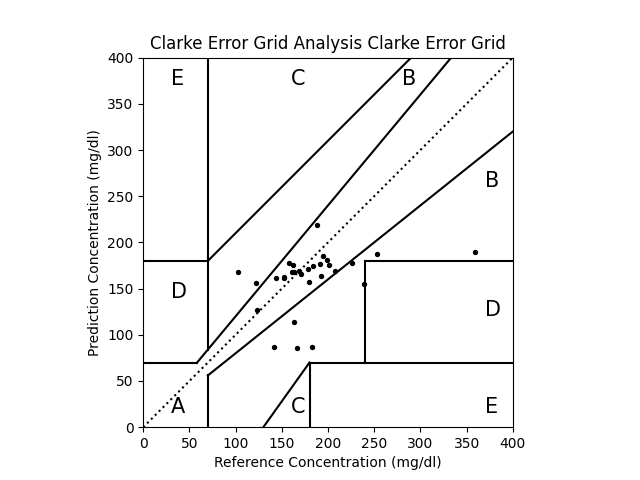}\label{subfig:tree-128-standard}}
    \hfill
    \subfigure[]{\includegraphics[width=0.49\textwidth]{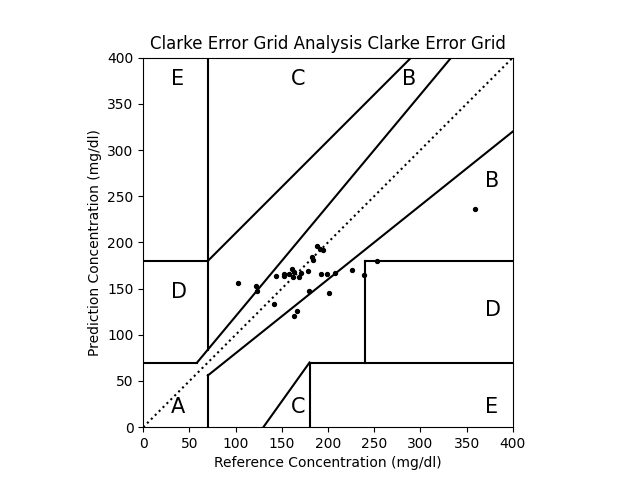}\label{subfig:tree-256-iit}}
    \subfigure[]{\includegraphics[width=0.49\textwidth]{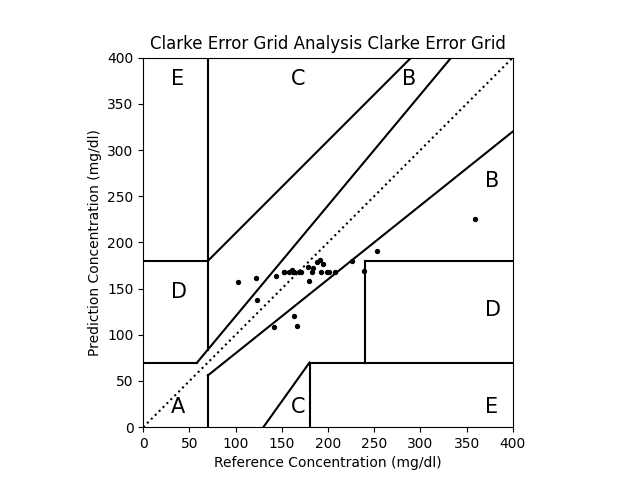}\label{subfig:tree-256-standard}}
    \caption{Error grid 120 PH of the MLP tree (128 hidden size) architecture (a) and (b), and  MLP tree (256 hidden size) architecture (c) and (d). The images on the left (a) and (c) refer to the models being trained through interchange intervention training using the regular causal model while the images on the right (b) and (d) correspond to conventional training; without IIT.}
\end{figure}

We used Clarke error grid analysis to visualize predictions within clinically accepted ranges. Only the 120 PH grid of predictions with random seed 56 is shown for each model. The model MLP parallel Figs.~\ref{subfig:parallel-iit},~\ref{subfig:parallel-standard} show a tendency to predict within the range of 140 and 180, nearly horizontal. On the other hand, the rest of the models Figs.~\ref{subfig:tree-128-iit},~\ref{subfig:tree-128-standard},~\ref{subfig:tree-256-iit},~\ref{subfig:tree-256-standard},~\ref{subfig:joint-iit},~\ref{subfig:joint-standard} show more dispersed predictions along the diagonal. This holds for both IIT-trained and standard-trained models.

\newpage

\begin{figure}[H]
    \centering
    \subfigure[]{\includegraphics[width=0.49\textwidth]{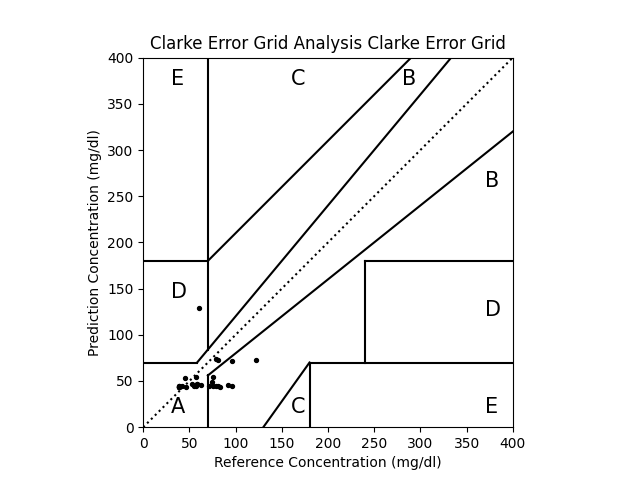}\label{subfig:tree-256-sim-iit}}
    \subfigure[]{\includegraphics[width=0.49\textwidth]{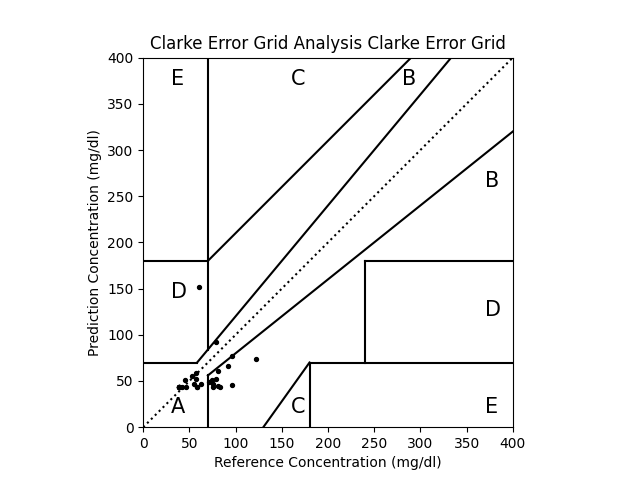}\label{subfig:tree-256-sim-standard}}
    \caption{Error grid 120 PH of the MLP tree (256 
 hidden size) architecture. The images on the left (a) refer to the model being trained through interchange intervention training using the amended causal model while the image on the right (b) correspond to conventional training; without IIT.}
\end{figure}
\end{document}